%% file: ms.tex
\documentclass[10pt,twocolumn,letterpaper]{article}

\usepackage{cvpr}
\usepackage{times}
\usepackage{epsfig}
\usepackage{graphicx}
\usepackage{amsmath}
\usepackage{amssymb}
\usepackage{url}
\usepackage{tabularx}
\usepackage{hhline}
\usepackage{multirow}
\usepackage{afterpage}
\usepackage{xcolor}
\usepackage{caption}
\usepackage{subcaption}
\usepackage{booktabs}
\usepackage{wrapfig}
\usepackage{lipsum}
\usepackage{titling}
\usepackage{stfloats}
\usepackage{enumitem}
\usepackage[flushleft]{threeparttable}
\usepackage{mathtools}
\usepackage{footmisc}
\usepackage{placeins}

\DeclareMathOperator*{\argmax}{arg\,max}
\DeclareMathOperator*{\argmin}{arg\,min}

\usepackage[pagebackref=true,breaklinks=true,letterpaper=true,colorlinks,bookmarks=false]{hyperref}

\cvprfinalcopy
\ifcvprfinal\fi
\newlength{\belowfigskip}
\makeatletter
\@namedef{ver@everyshi.sty}{}
\makeatother
\usepackage[absolute,showboxes]{textpos}
\TPMargin{4pt}
\newcommand{\IeeeCopyRightNotice}{
	\begin{textblock}{0.81}(0.08,0.912)
		\noindent \footnotesize \copyright 2019 IEEE. Personal use of this material is permitted. Permission from IEEE must be obtained for all other uses, in any current or future media, including reprinting/republishing this material for advertising or promotional purposes, creating new collective works, for resale or redistribution to servers or lists, or reuse of any copyrighted component of this work in other works.
		
		\vspace{2mm}
		
		\noindent \textbf{J\"org Wagner, Jan Mathias K\"ohler, Tobias Gindele, Leon Hetzel, Jakob Thadd\"aus Wiedemer, Sven Behnke}; The IEEE Conference on Computer Vision and Pattern Recognition (CVPR), 2019, pp. 9097-9107; the final, published version of this paper is available on IEEE Xplore.
	\end{textblock}
}

\begin{document}
\title{Interpretable and Fine-Grained Visual Explanations for \\Convolutional Neural Networks}
\date{}
\author{J\"org Wagner$^{1, 2}$ \hspace{7mm} Jan Mathias K\"ohler$^{1}$ \hspace{7mm} Tobias Gindele$^{1,}$\thanks{contributed while working at BCAI. We additionally thank Volker Fischer, Michael Herman, Anna Khoreva for discussions and feedback.} \hspace{7mm} Leon Hetzel$^{1,}$\footnotemark[1] \\ Jakob Thadd\"aus Wiedemer$^{1,}$\footnotemark[1]\ \hspace{7mm} Sven Behnke$^{2}$\\
$^1$Bosch Center for Artificial Intelligence (BCAI), Germany \hspace{6mm} $^2$University of Bonn, Germany\\
{\tt\small Joerg.Wagner3@de.bosch.com; behnke@cs.uni-bonn.de}
}

\maketitle
\thispagestyle{empty}
\vspace{1cm}

\IeeeCopyRightNotice

\input{abstract}
\input{introduction}
\input{related_work}
\input{methodology}
\input{experiments}

\input{conclusion}

\clearpage
\newpage
{\small
\bibliographystyle{ieee_fullname}
\bibliography{ms}
}

\newpage
\clearpage
\input{appendix}
\end{document}

%% file: abstract.tex
\begin{abstract}
To verify and validate networks, it is essential to gain insight into their decisions, limitations as well as possible shortcomings of training data. In this work, we propose a post-hoc, optimization based visual explanation method, which highlights the evidence in the input image for a specific prediction. Our approach is based on a novel technique to defend against adversarial evidence (\ie faulty evidence due to artefacts) by filtering gradients during optimization. The defense does not depend on human-tuned parameters. It enables explanations which are both fine-grained and preserve the characteristics of images, such as edges and colors. The explanations are interpretable, suited for visualizing detailed evidence and can be tested as they are valid model inputs. We qualitatively and quantitatively evaluate our approach on a multitude of models and datasets. 
\end{abstract}

%% file: introduction.tex
\section{Introduction}
Convolutional Neural Networks (CNNs) have proven to produce state-of-the-art results on a multitude of vision benchmarks, such as ImageNet~\cite{ILSVRC15}, Caltech~\cite{dollarCVPR09peds} or Cityscapes~\cite{cordts2016cityscapes} which led to CNNs being used in numerous real-world systems (\eg autonomous vehicles) and services (\eg translation services). Though, the use of CNNs in safety-critical domains presents engineers with challenges resulting from their black-box character. A better understanding of the inner workings of a model provides hints for improving it, understanding failure cases and it may reveal shortcomings of the training data. Additionally, users generally trust a model more when they understand its decision process and are able to anticipate or verify outputs~\cite{mcallister2017concrete}.
\begin{figure}[t!]
	\includegraphics[width=\linewidth]{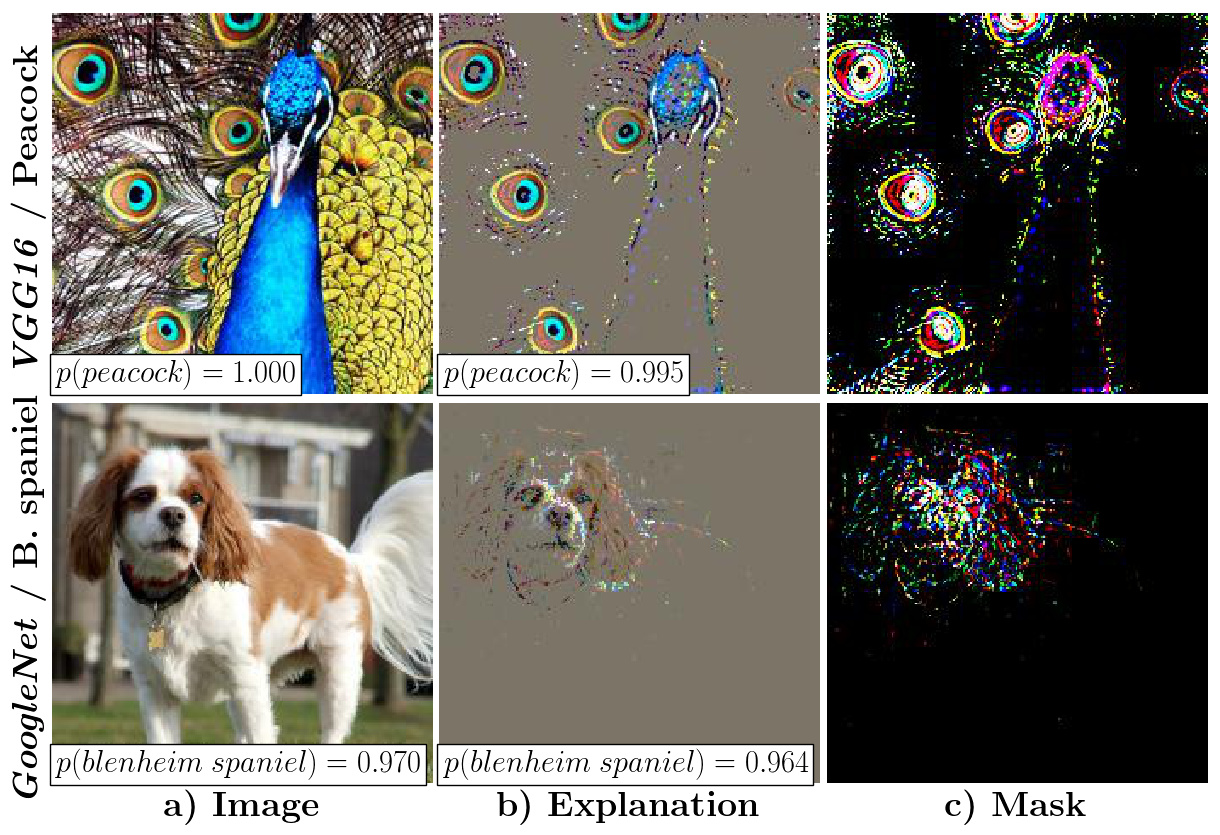}
	\vspace{-0.6cm}
	\caption{Fine-grained explanations computed by removing irrelevant pixels. a) Input image with softmax score $p(c_{ml})$ of the most-likely class. Our method tries to find a sparse mask (c) with irrelevant pixels set to zero. The resulting explanation (b), \ie: 'image\,$\times$\,mask', is optimized in the image space and, thus, can directly be used as model input. The parameter $\lambda$ is optimized to produce an explanation with a softmax score comparable to the image.}
	\label{fig:vis_explainations_intro}
	\vspace{\belowfigskip}
\end{figure}
 
To overcome the interpretation and transparency disadvantage of black-box models, post-hoc explanation methods have been introduced~\cite{zhou2016learning, selvaraju2017grad, springenberg2015striving, zhang2016top, petsiuk2018rise, fong2017interpretable, dabkowski2017real}. These methods provide explanations for individual predictions and thus help to understand on which evidence a model bases its decisions. The most common form of explanations are visual, image-like representations, which depict the important pixels or image regions in a human interpretable manner. 

In general, an explanation should be easily interpretable (Sec.~\ref{sec:interpretability}). Additionally, a visual explanation should be class discriminative and fine-grained~\cite{selvaraju2017grad} (Sec.~\ref{sec:class_disc}). The latter property is particularly important for classification tasks in the medical~\cite{gulshan2016development, gondal2017weakly} domain, where fine structures (\eg capillary hemorrhages) have a major influence on the classification result (Sec.~\ref{sec:retina}). Besides, the importance of different color channels should be captured, \eg to uncover a color bias in the training data (Sec.~\ref{sec:biases}). 

Moreover, explanations should be faithful, meaning they accurately explain the function of the black-box model~\cite{selvaraju2017grad}. To evaluate the faithfulness (Sec.~\ref{sec:faithfulness}), recent work~\cite{selvaraju2017grad, petsiuk2018rise, chattopadhyay2017grad} introduce metrics which are based on model predictions of explanations. To be able to compute such metrics without having to rely on proxy measures~\cite{selvaraju2017grad}, it is beneficial to employ explanation methods which directly generate valid model inputs (\eg a perturbed version of the image).

A major concern of optimization based visual explanation methods is adversarial evidence, \ie faulty evidence generated by artefacts introduced in the computation of the explanation. Therefore, additional constraints or regularizations are used to prevent such faulty evidence~\cite{fong2017interpretable, dabkowski2017real, du2018towards}. A drawback of these defenses are added hyperparameters and the necessity of either a reduced resolution of the explanation or a smoothed explanation (Sec.~\ref{sec:intrinsic_defense}), thus, they are not well suited for displaying fine-grained evidence. 

Our main contribution is a new adversarial defense technique which selectively filters gradients in the optimization which would lead to adversarial evidence otherwise (Sec.~\ref{sec:intrinsic_defense}). Using this defense, we extend the work of~\cite{fong2017interpretable} and propose a new fine-grained visual explanation method (FGVis). The proposed defense is not dependend on hyperparameters and is the key to produce fine-grained explanations (Fig.~\ref{fig:vis_explainations_intro}) as no smoothing or regularizations are necessary. Like other optimization-based approaches, FGVis computes a perturbed version of the original image, in which either all irrelevant or the most relevant pixels are removed. The resulting explanations  (Fig~\ref{fig:vis_explainations_intro}\,b) are valid model inputs and their faithfulness can, thus, be directly verified (as in methods from~\cite{fong2017interpretable, du2018towards, chang2018explaining, dabkowski2017real}). Moreover, they are additionally fine-grained (as in methods from~\cite{selvaraju2017grad, simonyan2013deep, zeiler2014visualizing, springenberg2015striving}). To the best of our knowledge, this is the first method to be able to produce fine-grained explanations directly in the image space. We evaluate our defense (Sec.~\ref{sec:intrinsic_defense}) and FGVis (Sec.~\ref{sec:qualitative_results} and~\ref{sec:quantitative_results}) qualitatively and quantitatively.

%% file: related_work.tex
\section{Related Work}
\label{sec:related_work}
Various methods to create explanations have been introduced. Thang~\etal~\cite{zhang2018visual} and DU~\etal~\cite{du2018techniques} provide a survey of these. In this section, we give an overview of explanation methods which generate visual, image-like explanations. 

\noindent\textbf{Backpropagation Based Methods (BBM).} These methods generate an importance measure for each pixel by backpropagating an error signal to the image. Simonyan~\etal~\cite{simonyan2013deep}, which build on work of Baehrens~\etal~\cite{baehrens2010explain}, use the derivative of a class score with respect to the image as an importance measure. Similar methods have been introduced in Zeiler~\etal~\cite{zeiler2014visualizing} and Springenberg~\etal~\cite{springenberg2015striving}, which additionally manipulate the gradient when backpropagating through ReLU nonlinearities. Integrated Gradients~\cite{sundararajan2017axiomatic} additionally accumulates gradients along a path from a base image to the input image. SmoothGrad~\cite{smilkov2017smoothgrad} and VarGrad~\cite{adebayo2018local} visually sharpen explanations by combining multiple explanations of noisy copies of the image. Other BBMs such as Layer-wise Relevance Propagation~\cite{bach2015pixel}, DeepLift~\cite{shrikumar2017learning} or Excitation Backprop~\cite{zhang2016top} utilize top-down relevancy propagation rules. BBMs are usually fast to compute and produce fine-grained importance/relevancy maps. However, these maps are generally of low quality~\cite{dabkowski2017real, du2018towards} and are less interpretable. To verify their faithfulness it is necessary to apply proxy measures or use pre-processing steps, which may falsify the result. 

\noindent\textbf{Activation Based Methods (ABM).} These approaches use a linear combination of activations from convolutional layers to form an explanation. Prominent methods of this category are CAM (Class Activation Mapping)~\cite{zhou2016learning} and its generalizations Grad-CAM~\cite{selvaraju2017grad} and Grad-CAM++~\cite{chattopadhyay2017grad}. These methods mainly differ in how they calculate the weights of the linear combination and what restrictions they impose on the CNN. Extensions of such approaches have been proposed in Selvaraju~\etal~\cite{selvaraju2017grad} and Du~\etal~\cite{du2018towards}, which combine ABMs with backpropagation or perturbation based approaches. ABMs generate easy to interpret heat-maps which can be overlaid on the image. However, they are generally not well suited to visualize fine-grained evidence or color dependencies. Additionally, it is not guaranteed that the resulting explanations are faithful and reflect the decision making process of the model~\cite{du2018towards, selvaraju2017grad}.

\noindent\textbf{Perturbation Based Methods (PBM).} Such approaches perturb the input and monitor the prediction of the model. Zeiler~\etal~\cite{zeiler2014visualizing} slide a grey square over the image and use the change in class probability as a measure of importance. Several approaches are based on this idea, but use other importance measures or occlusion strategies. Petsiuk~\etal~\cite{petsiuk2018rise} use randomly sampled occlusion masks and define importance based on the expected model score over masks. LIME~\cite{ribeiro2016should} uses a super-pixel based occlusion strategy and a surrogate model to compute importance scores. Further super-pixel or segment based methods are introduced in Seo~\etal~\cite{seo2018regional} and Zhou~\etal~\cite{zhou2014object}. The so far mentioned approaches do not need access to the internal state or structure of the model. Though, they are often quite time consuming and only generate coarse explanations. 
  
Other PBMs generate an explanation by optimizing for a perturbed version of the image~\cite{dabkowski2017real, fong2017interpretable, du2018towards, chang2018explaining}. The perturbed image $\mathbf{e}$ is defined by $\mathbf{e} = \mathbf{m} \cdot \mathbf{x} + (1-\mathbf{m}) \cdot \mathbf{r}$, where $\mathbf{m}$ is a mask, $\mathbf{x}$ the input image, and $\mathbf{r}$ a reference image containing little information (Sec.~\ref{sec:perturbation_based_explanations}). To avoid adversarial evidence, these approaches need additional regularizations~\cite{fong2017interpretable}, constrain the explanation (\eg optimize for a coarse mask~\cite{chang2018explaining, fong2017interpretable, du2018towards}), introduce stochasticity~\cite{fong2017interpretable}, or utilize regularizing surrogate models~\cite{dabkowski2017real}. These approaches generate easy to interpret explanations in the image space, which are valid model inputs and faithful (\ie a faithfulness measure is incorporated in the optimization). 

Our method also optimizes for a perturbed version of the input. Compared to existing approaches we propose a new adversarial defense technique which filters gradients during optimization. This defense does not need hyperparameters which have to be fine-tuned. Besides, we optimize each pixel individually, thus, the resulting explanations have no limitations on the resolution and are fine-grained. 

%% file: methodology.tex
\section{Explaining Model Predictions}  
\label{sec:approach}
Explanations provide insights into the decision-making process of a model. The most universal form of explanations are \textit{global} ones which characterize the overall model behavior. \textit{Global} explanations specify for all possible model inputs the corresponding output in an intuitive manner.  A decision boundary plot of a classifier in a low-dimensional vector space, for example, represents a \textit{global} explanation. For high-dimensional data and complex models, it is practically impossible to generate such explanations. Current approaches therefore utilize \textit{local} explanations\footnote{For the sake of brevity, we will use the term explanations as a synonym for \textit{local} explanations throughout this work.}, which focus on individual inputs. Given one data point, these methods highlight the evidence on which a model bases its decisions. As outlined in Sec.~\ref{sec:related_work}, the definition of highlighting depends on the used explanation method. In this work, we follow the paradigm introduced in \cite{fong2017interpretable} and directly optimize for a perturbed version of the input image. Such an approach has several advantages: 1) The resulting explanations are interpretable due to their image-like nature; 2) Explanations represent valid model inputs and are thus testable; 3) Explanations are optimized to be faithful. In Sec.~\ref{sec:perturbation_based_explanations} we briefly review the general paradigm of optimization based explanation methods before we introduce our novel adversarial defense technique in Sec.~\ref{sec:intrinsic_defense}.

\subsection{Perturbation based Visual Explanations}
\label{sec:perturbation_based_explanations}
Following the paradigm of optimization based explanation methods, which compute a perturbed version of the image~\cite{fong2017interpretable, du2018towards, chang2018explaining, dabkowski2017real}, an explanation can be defined as:\\
\noindent\textbf{Explanation by Preservation:} The smallest region of the image which must be retained to preserve the original model output (\ie minimal sufficient evidence).\\
\noindent\textbf{Explanation by Deletion:} The smallest region of the image which must be deleted to change the model output.

To formally derive an explanation method based on this paradigm, we assume that a CNN $f_{cnn}$ is given which maps an input image $\mathbf{x} \in \mathbb{R}^{3 \times H \times W}$ to an output $\mathbf{y}_{x}=f_{cnn}(\mathbf{x}; \theta_{cnn})$. The ouput $\mathbf{y}_{x} \in \mathbb{R}^{C}$ is a vector representing the softmax scores  $y_{x}^{c}$ of the different classes $c$. Given an input image $\mathbf{x}$, an explanation $\mathbf{e}^{\ast}_{c_T}$ of a target class $c_{T}$ (\eg the most-likely class $c_{T}=c_{ml}$) is computed by removing either relevant (\textit{deletion}) or irrelevant, not supporting $c_{T}$, information (\textit{preservation}) from the image. Since it is not possible to remove information without replacing it, and we do not have access to the image generating process, we have to use an approximate removal operator~\cite{fong2017interpretable}. A common approach is to use a mask based operator $\Phi$, which computes a weighted average between the image $\mathbf{x}$ and a reference image $\mathbf{r}$, using a mask $\mathbf{m}_{c_T} \in \left[ 0,1\right] ^{3 \times H \times W}$: 
\begin{equation}
	\mathbf{e}_{c_T} = \Phi(\mathbf{x}, \mathbf{m}_{c_T}) = \mathbf{x} \cdot \mathbf{m}_{c_T} + (1 - \mathbf{m}_{c_T}) \cdot \mathbf{r}.
	\label{eq:mask_operator}
\end{equation}
Common choices for the reference image are constant values (\eg zero), a blurred version of the original image, Gaussian noise, or sampled references of a generative model~\cite{fong2017interpretable, du2018towards, chang2018explaining, dabkowski2017real}. In this work, we take a zero image as reference. In our opinion, this reference produces the most pleasing visual explanations, since irrelevant image areas are set to zero\footnote{Tensors $\mathbf{x, e, r}$ are assumed to be normalized according to the training of the CNN. A value of zero for these thus corresponds to a grey color (\ie the color of the data mean).} (Fig.~\ref{fig:vis_explainations_intro}) and not replaced by other structures. In addition, the zero image (and random image) carry comparatively little information and lead to a model prediction with a high entropy. Other references, such as a blurred version of the image, usually result in lower prediction entropies, as shown in Sec.~\ref{sec:appx_references}. Due to the additional computational effort, we have not considered model-based references as proposed in Chang~\etal~\cite{chang2018explaining}. 

In addition, a similarity metric $\varphi(y^{c_{T}}_{x}, y^{c_{T}}_{e})$ is needed, which measures the consistency of the model output generated by the explanation $y^{c_{T}}_{e}$ and the output of the image $y^{c_{T}}_{x}$ with respect to a target class $c_{T}$. This similarity metric should be small if the explanation preserves the output of the target class and large if the explanation manages to significantly drop the probability of the target class~\cite{fong2017interpretable}. Typical choices for the metric are the cross-entropy with the class $c_T$ as a hard target~\cite{hinton2015distilling} or the negative softmax score of the target class $c_T$. The similarity metric ensures that the explanation remains faithful to the model and thus accurately explains the function of the model, this property is a major advantage of PBMs. 

\begin{figure*}[t!]
	\centering	 
	\includegraphics[width=0.92\linewidth]{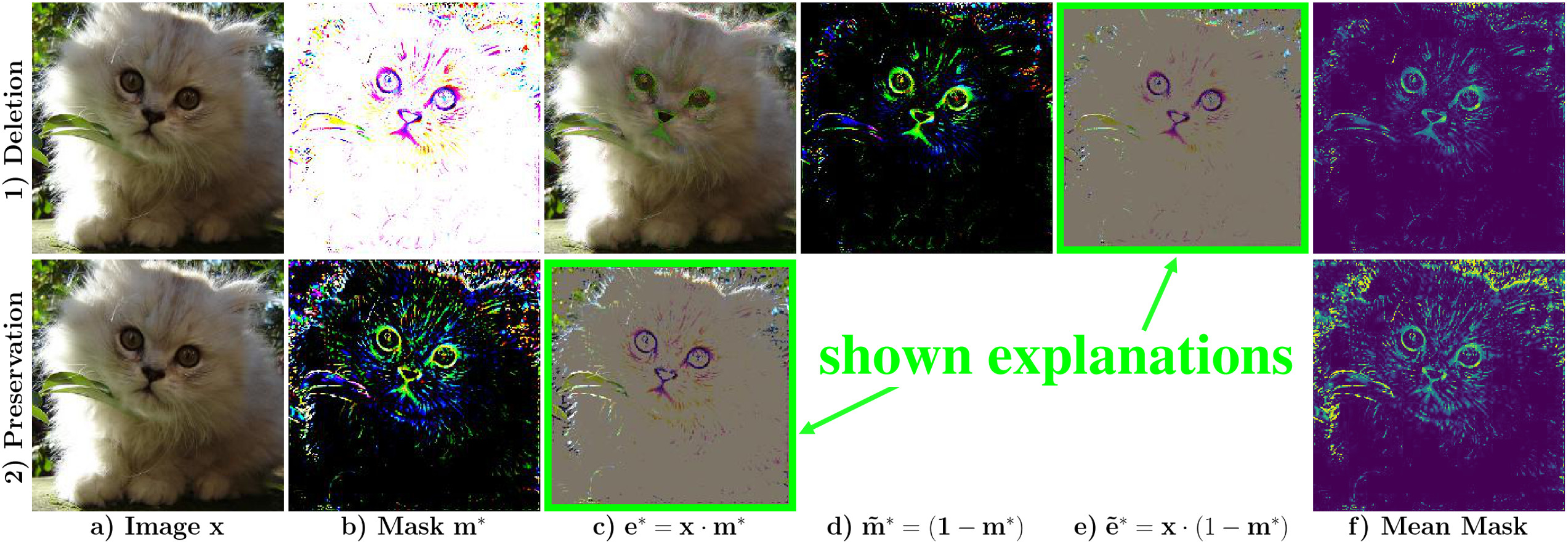}
	\vspace{-0.15cm}
	\caption{Visualization types calculated for \textit{VGG} using \textit{deletion} / \textit{preservation game}. For the \textit{repression} / \textit{generation game} the same characteristics hold. Subscript $c_T$ ommited to ease readability. a) Input image. b) Mask obtained by the optimization. Colors in a \textit{deletion} mask are complementary to the image colors. c) Explanation directly obtained by the optimization. d) Complementary mask with a true-color representation for the \textit{deletion game}. e)~Explanation highlighting the important evidence for the \textit{deletion game}. f) Mean mask: mask / comp. mask averaged over colors. --- 
	To underline important evidence, we use $\mathbf{e}$ for the explanation of the \textit{preservation\;/\;generation game} and $\mathbf{\tilde{e}}$ for the \textit{deletion\;/\;repression game}.}
	\label{fig:vis_types}
	\vspace{\belowfigskip}
\end{figure*}

Using the mask based definition of an explanation with a zero image as reference ($\mathbf{r}=\mathbf{0}$) as well as the similarity metric, a \textit{preserving explanation} can be computed by:
\begin{equation}
	\begin{aligned}
		\mathbf{e}^{\ast}_{c_T} &= \mathbf{m}^{\ast}_{c_T} \cdot \mathbf{x}, \\
		\mathbf{m}^{\ast}_{c_T} &= \argmin_{\mathbf{m}_{c_T}} \{ \varphi(y^{c_{T}}_{x}, y^{c_{T}}_{e}) + \lambda \cdot \left\| \mathbf{m}_{c_T} \right\|_{1} \}.
		\label{eq:preservation_game}
	\end{aligned}
\end{equation}
We will refer to the optimization in Eq.~\ref{eq:preservation_game} as the \textit{preservation game}. Masks (Fig.~\ref{fig:vis_types}\textcolor{red}{\,/\,b2})\footnote{Fig.~\ref{fig:vis_types}\textcolor{red}{\,/\,b2}: Figure~\ref{fig:vis_types}, column b, 2nd row} generated by this game are sparse (\ie many pixels are zero / appear black; enforced by minimizing $\left\| \mathbf{m}_{c_T} \right\|_{1}$) and only contain large values at most important pixels. The corresponding explanation is computed by multiplying the mask with the image (Fig.~\ref{fig:vis_types}\textcolor{red}{\,/\,c2}). 

\noindent Alternatively, we can compute a \textit{deleting explanation} using:
\begin{equation}
	\begin{aligned}
		\mathbf{e}^{\ast}_{c_T} &= \mathbf{m}^{\ast}_{c_T} \cdot \mathbf{x}, \\
		\mathbf{m}^{\ast}_{c_T} &= \argmax_{\mathbf{m}_{c_T}} \{\varphi(y^{c_{T}}_{x}, y^{c_{T}}_{e}) + \lambda \cdot \left\| \mathbf{m}_{c_T} \right\|_{1} \}.
		\label{eq:deletion_game}
	\end{aligned}
\end{equation}
This optimization will be called \textit{deletion game} henceforward. Masks (Fig.~\ref{fig:vis_types}\textcolor{red}{\,/\,b1}) generated by this game contain mainly ones (\ie appear white; enforced by maximizing $\left\| \mathbf{m}_{c_T} \right\|_{1}$ in Eq.~\ref{eq:deletion_game}) and only small entries at pixels, which provide the most prominent evidence for the target class. The colors in a mask of the \textit{deletion game} are complementary to the image colors. To obtain a true-color representation analogous to the \textit{preservation game}, one can alternatively visualize the complementary mask (Fig.~\ref{fig:vis_types}\textcolor{red}{\,/\,d1}): $\mathbf{\tilde{m}}^{\ast}_{c_T}=(\mathbf{1}-\mathbf{m}^{\ast}_{c_T})$. A resulting explanation of the \textit{deletion game}, as defined in Eq.~\ref{eq:deletion_game}, is visualized in Fig.~\ref{fig:vis_types}\textcolor{red}{\,/\,c1}. This explanation is visually very similar to the original image as only a few pixels need to be deleted to change the model output. In the remaining of the paper for better visualization, we depict a modified version of the explanation for the \textit{deletion game}: $\mathbf{\tilde{e}}^{\ast}_{c_T} = \mathbf{x} \cdot (\mathbf{1} - \mathbf{m}^{\ast}_{c_T})$. This explanation has the same properties as the one of the \textit{preservation game}, \ie it only highlights the important evidence. We observe that the \textit{deletion game} generally produces sparser explanations compared to the \textit{preservation game}, as less pixels have to be removed to delete evidence for a class than to maintain evidence by preserving pixels.

To solve the optimization in Eq.~\ref{eq:preservation_game} and Eq.~\ref{eq:deletion_game}, we utilize Stochastic Gradient Descent and start with an explanation $\mathbf{e}_{c_T}^{0}=\mathbf{1} \cdot \mathbf{x}$ identical to the original image (\ie a mask initialized with ones). As an alternative initialization of the masks, we additionally explore a zero initialization $\mathbf{m}_{c_T}^{0}=\mathbf{0}$. 
In this setting the initial explanation contains no evidence towards any class and the optimization iteratively has to add relevant (\textit{generation game}) or irrelevant, not supporting the class $c_T$, information (\textit{repression game}). The visualizations of the \textit{generation game} are equivalent to those of the \textit{preservation game}, the same holds for the \textit{deletion} and \textit{repression game}. In our experiments the \textit{deletion game} produces the most fine-grained and visually pleasing explanations. Compared to the other games it usually needs the least amount of optimization iterations since we start with $\mathbf{m}_{c_T}^{0}=\mathbf{1}$ and comparatively few mask values have to be changed to delete the evidence for the target class. A comparison and additional characteristics of the four optimization settings (\ie games) are included in Sec.~\ref{sec:appx_comparisonGames}.

\subsection{Defending against Adversarial Evidence}
\label{sec:intrinsic_defense}
CNNs have been proven susceptible to adversarial images \cite{Szegedy2014Intriguing, Goodfellow2015Explaining, kurakin2016adversarial}, i.e. a perturbed version of a correctly classified image crafted to fool a CNN. Due to the computational similarity of adversarial methods and optimization based visual explanation approaches, adversarial noise is also a concern for the latter methods and one has to ensure that an explanation is based on true evidence present in the image and not on false adversarial evidence introduced during optimization. This is particularly true for the \textit{generation/repression game} as their optimization start with $\mathbf{m}_{c_T}^{0}=\mathbf{0}$ and iteratively adds information.

\cite{fong2017interpretable} and~\cite{dabkowski2017real} showed the vulnerability of optimization based explanation methods to adversarial noise. To avoid adversarial evidence, explanation methods use stochastic operations~\cite{fong2017interpretable}, additional regularizations~\cite{fong2017interpretable, dabkowski2017real}, optimize on a low-resolution mask with upsampling of the computed mask~\cite{fong2017interpretable, du2018towards, chang2018explaining}, or utilize a regularizing surrogate model~\cite{dabkowski2017real}. In general, these operations impede the generation of adversarial noise by obscuring the gradient direction in which the model is susceptible to false evidence, or by constraining the search space for potential adversarials. These techniques help to reduce adversarial evidence, but also introduce new drawbacks: 1)~Defense capabilities usually depend on human-tuned parameters; 2)~Explanations are limited to being low resolution and/or smooth, which prevents fine-grained evidence from being visualized. 

\begin{figure}[t]
	\begin{center}
		\includegraphics[width=\linewidth]{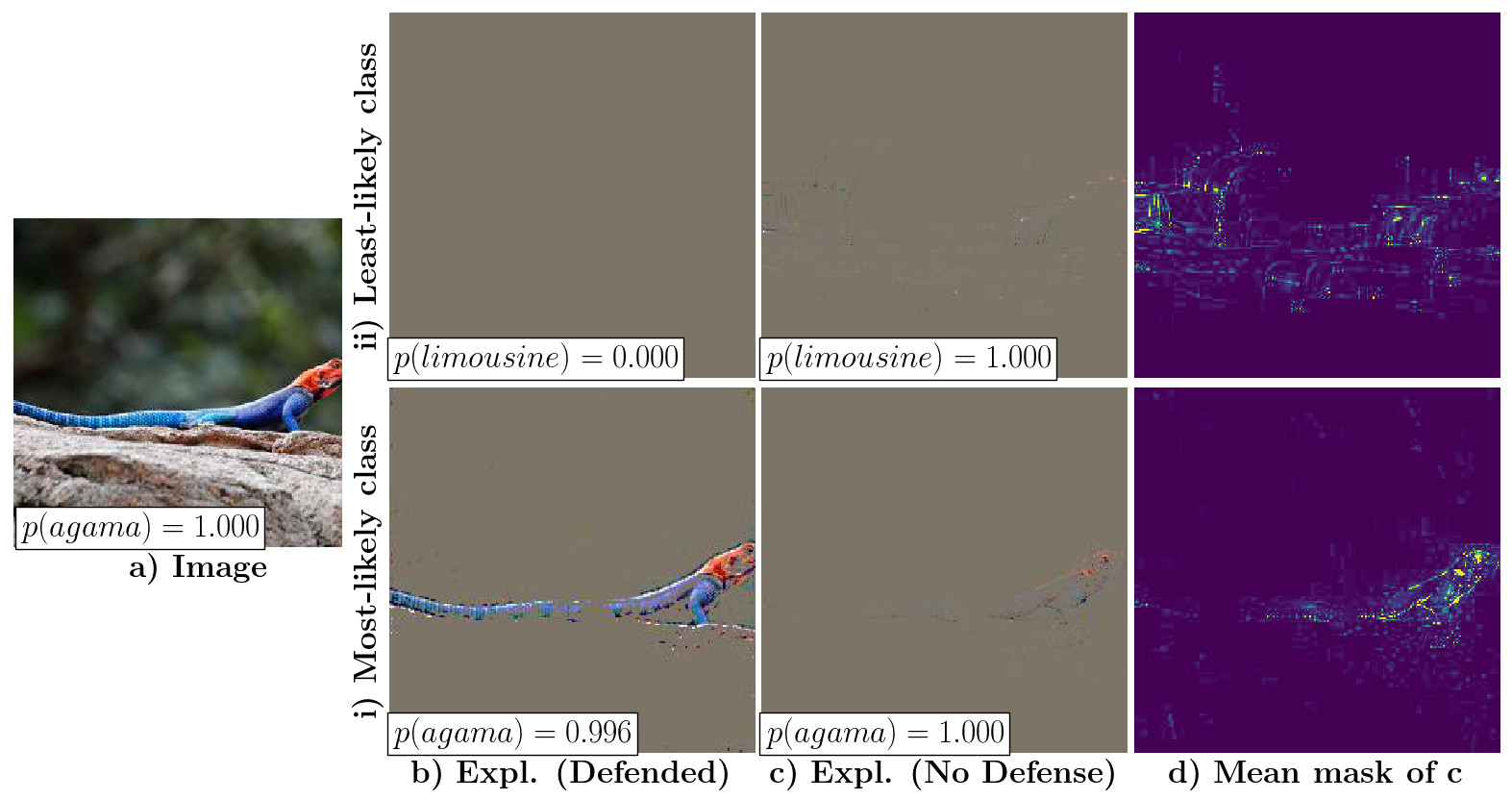}
	\end{center}
	\vspace{-0.5cm}
	\caption{Explanations computed for the adversarial class \textit{limousine} and the predicted class \textit{agama} using the \textit{generation game} and \textit{VGG16} with and without our adversarial defense. An adversarial for class \textit{limousine} can only be computed without the defense. d)~Mean mask enhanced by a factor of $7$ to show small adversarial structures. }
	\label{fig:adversarial_evidence}
	\vspace{\belowfigskip}
\end{figure}
\indent\textbf{A novel Adversarial Defense.} 
To overcome these drawbacks, we propose a novel adversarial defense which filters gradients during backpropagation in a targeted way. The basic idea of our approach is: A neuron within a CNN is only allowed to be activated by the explanation $\mathbf{e}_{c_T}$ if the same neuron was also activated by the original image $\mathbf{x}$. If we regard neurons as indicators for the existence of features (\eg edges, object parts, \dots), the proposed constraint enforces that the explanation $\mathbf{e}_{c_T}$ can only contain features which exist at the same location in the original image $\mathbf{x}$. By ensuring that the allowed features in $\mathbf{e}_{c_T}$ are a subset of the features in $\mathbf{x}$ it prevents the generation of new evidence.

This defense technique can be integrated in the introduced explanation methods via an optimization constraint:
\begin{equation}
	\begin{cases}
		0 \leq h^{l}_{i}(\mathbf{e}_{c_T}) \leq h^{l}_{i}(\mathbf{x}), & \text{if }\, h^{l}_{i}(\mathbf{x}) \geq 0, \\
		0  \geq h^{l}_{i}(\mathbf{e}_{c_T}) \geq h_{i}^{l}(\mathbf{x}), & \text{otherwise}, \\
	\end{cases}
	\label{eq:constraint}
\end{equation}
where $h^{l}_{i}$ is the activation of the $i$-th neuron  in the $l$-th layer of the network after the nonlinearity. For brevity, the index $i$ references one specific feature at one spatial position in the activation map. This constraint is applied after all nonlinearity-layers (\eg ReLU-Layers) of the network, besides the final classification layer. It ensures that the absolute value of activations can only be reduced towards values representing lower information content (we assume that zero activations have the lowest information as commonly applied in network pruning~\cite{han2015learning}). To solve the optimization with subject to Eq.~\ref{eq:constraint}, one could incorporate the constraints via a penalty function in the optimization loss. The drawback is one additional hyperparameter. Alternatively, one could add an additional layer $\bar{h}^{l}_{i}$ after each nonlinearity which ensures the validity of Eq.~\ref{eq:constraint}:
\begin{equation}
	\begin{aligned}
		\bar{h}^{l}_{i}(\mathbf{e}_{c_T}) &= \min (bu, \max (bl, {h}^{l}_{i}(\mathbf{e}_{c_T}) ) ), \\
		bu &= \max (0, h^{l}_{i}(\mathbf{x})), \\
		bl &= \min (0, h^{l}_{i}(\mathbf{x})),  			
		\label{eq:activation_clipping}
	\end{aligned}
\end{equation}
where ${h}^{l}_{i}(\mathbf{e}_{c_T})$ is the actual activation of the original nonlinearity-layer and $\bar{h}^{l}_{i}(\mathbf{e}_{c_T})$ the adjusted activation after ensuring the bounds $bu$, $bl$ of the original input. For instance, for a ReLU nonlinearity, the upper bound $bu$ is equal to $h^{l}_{i}(\mathbf{x})$ and the lower bound $bl$ is zero. We are not applying this method as it changes the architecture of the model which we try to explain. Instead, we clip gradients in the backward pass of the optimization, which lead to a violation of Eq.~\ref{eq:constraint}. This is equivalent to adding an additional clipping-layer after each nonlinearity which acts as the identity in the forward pass and uses the gradient update of Eq.~\ref{eq:activation_clipping} in the backward pass. When backpropagating an error-signal $\bar{\gamma}^{l}_{i}$ through the clipping-layer, the gradient update rule for the resulting error $\gamma^{l}_{i}$ is defined by:
\begin{equation} 
	\gamma^{l}_{i} = \bar{\gamma}^{l}_{i} \cdot [{h}^{l}_{i}(\mathbf{e}_{c_T}) \leq bu] \cdot [{h}^{l}_{i}(\mathbf{e}_{c_T}) \geq bl],
	\label{eq:gradient_update_rule}
\end{equation}
where $[\,\cdot\,]$ is the indicator function and $bl$, $bu$ the bounds computed in Eq.~\ref{eq:activation_clipping}. This clipping only affects the gradients of the similarity metric $\varphi(\cdot\, , \cdot)$ which are propagated through the network. The proposed gradient clipping does not add hyperparameters and keeps the original structure of the model during the forward pass. Compared to other adversarial defense techniques (\cite{dabkowski2017real}, \cite{fong2017interpretable}, \cite{chang2018explaining}), it imposes no constraint on the explanation (\eg resolution/smoothness constraints), enabling fine-grained explanations. \\
\indent\textbf{Validating the Adversarial Defense.}
To evaluate the performance of our defense, we compute an explanation for a class $c_A$ for which there is no evidence in the image (\ie it is visually not present). We approximate $c_A$ with the least-likely class $c_{ll}$ considering only images which yield very high predictive confidence for the true class $p(c_{true}) \geq 0.995$. Using $c_{ll}$ as the target class, the resulting explanation method without defense is similar to an adversarial attack (the \textit{Iterative Least-Likely Class Method}~\cite{kurakin2016adversarial}).

A correct explanation for the adversarial class $c_A$ should be ``empty'' (\ie grey), as seen in Fig.~\ref{fig:adversarial_evidence}\,b, top row, when using our adversarial defense. If, on the other hand, the explanation method is susceptible to adversarial noise, the optimization procedure should be able to perfectly generate an explanation for any class. This behavior can be seen in Fig.~\ref{fig:adversarial_evidence}\,c, top row. The shown explanation for the adversarial class ($c_A$:~\textit{limousine}) contains primarily artificial structures and is classified with a probability of $1$ as \textit{limousine}. 

We also depict the explanation of the predicted class ($c_{pred}$:~\textit{agama}). The explanation with our defense results in a meaningful representation of the \textit{agama} (Fig.~\ref{fig:adversarial_evidence}\,b, bottom row); without defense (Fig.~\ref{fig:adversarial_evidence}\,c\,/\,d, bottom row) it is much more sparse. As there is no constraint to change pixel values arbitrarily, we assume the algorithm introduces additional structures to produce a sparse explanation. 

\begin{table}[t!]
	\centering
	\scalebox{0.9}{
		\begin{tabular}{ccc}
			\hline
			Model 												& No Defense 			& Defended \\
			\hline 
			\textit{VGG16}~\cite{simonyan2014very} 				& $100.0\,\%$ 			& $0.2\,\%$ \\ 
			\textit{AlexNet}~\cite{krizhevsky2012imagenet} 		& $100.0\,\%$ 			& $0.0\,\%$ \\
			\textit{ResNet50}~\cite{He16} 						& $100.0\,\%$ 			& $0.0\,\%$ \\
			\textit{GoogleNet}~\cite{szegedy2015going} 			& $100.0\,\%$ 			& $0.0\,\%$ \\ 
			\hline
		\end{tabular}
	}
	\caption{Ratio how often an adversarial class $c_{A}$ was generated, using the \textit{generation game} with no sparsity loss on \textit{VGG16} with and without our defense.}
	\label{table:adversarial_defense}
	\vspace{\belowfigskip}
\end{table}
A quantitative evaluation of the proposed defense is reported in Tab.~\ref{table:adversarial_defense}. We generate explanations for 1000 random ImageNet validation images and use a class $c_A$ as the explanation target\footnote{For $c_A$ we used the least-likely class, as described before. We use the second least-likely class, if the least-likely class coincidentally matches the predicted class for the zero image.}. To ease the generation of adversarial examples, we set the sparsity loss to zero and only use the similarity metric which tries to maximize the probability of the target class $c_{A}$. Without an employed defense technique, the optimization is able to generate an adversarial explanation for 100\% of the images. Applying our defense (Eq.~\ref{eq:gradient_update_rule}), the optimization nearly never was able to do so. The two adverarial examples generated in \textit{VGG16} have a low confidence, so we assume that there has been some evidence for the chosen class $c_{A}$ in the image. Our proposed technique is thus well suited to defend against adversarial evidence. 

%% file: experiments.tex
\section{Qualitative Results} \label{sec:qualitative_results}
Implementation details are stated in~Sec.~\ref{sec:appx_implementation}.
\input{interpretability}
\input{class_discrimination}
\input{bias_training_data}

\section{Quantitative Results} \label{sec:quantitative_results}
\input{faithfulness_explanations}
\input{retina}

%% file: interpretability.tex
\subsection{Interpretability}
\label{sec:interpretability}

\textbf{Comparison of methods.} Using the \textit{deletion game} we compute mean explanation masks for \textit{GoogleNet} and compare these in~Fig.~\ref{fig:comparison} with state-of-the-art methods. Our method delivers the most fine-grained explanation by deleting important pixels of the target object. Especially explanations b), f), and g) are coarser and, therefore, tend to include background information not necessary to be deleted to change the original prediction. The majority of pixels highlighted by FGVis form edges of the object. This cannot be seen in other methods. The explanations from c) and d) are most similar to ours. However, our masks are computed to directly produce explanations which are viable network inputs and are, therefore, verifiable --- The deletion of the highlighted pixels prevents the model from correctly predicting the object. This statement does not necessarily hold for explanations calculated with methods c) and d).\\ 
\indent\textbf{Architectural insights.} As first noted in~\cite{nie2018theoretical} explanations using backpropagation based approaches show a grid-like pattern for ResNet. In general, \cite{nie2018theoretical} demonstrate that the network structure influences the visualization and assume that for ResNet the skip connections play an important role in their explanation behavior. As shown in Fig~\ref{fig:resnet_grid_pattern} this pattern is also visible in our explanations to an even finer degree. Interestingly, the grid pattern is also visible to a lesser extent outside the object. A detailed investigation of this phenomenon is left for future research. See~\ref{sec:appx_comparisonNets} for a comparison of explanations between models.

%% file: class_discrimination.tex
\subsection{Class Discriminative / Fine-Grained}
\label{sec:class_disc}
\begin{figure}[b!]
	\centering
	\includegraphics[width=0.8\linewidth]{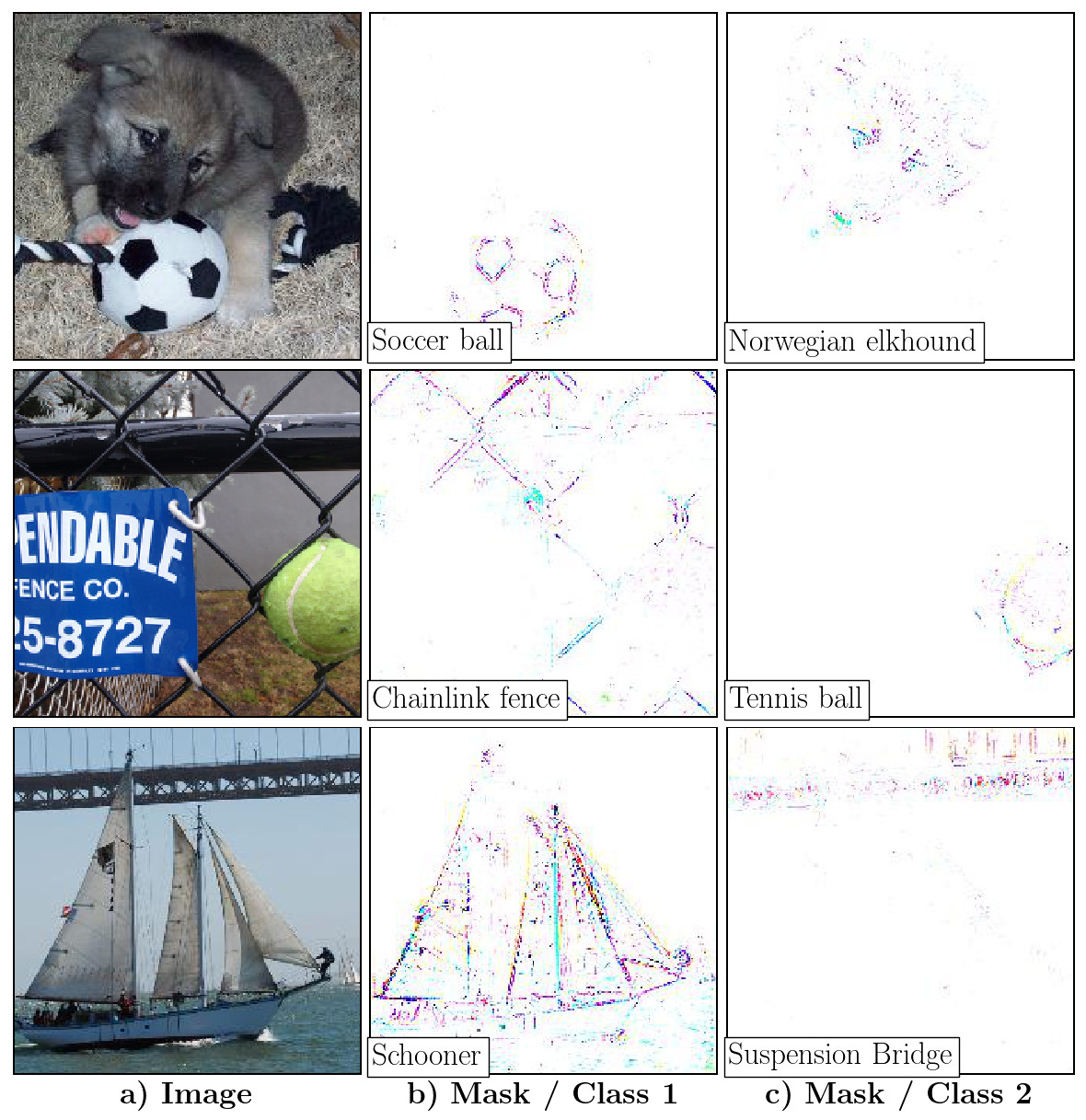}
	\caption{Explanation masks for images with multiple objects computed using the \textit{deletion game} and \textit{GoogleNet}. FGVis produces class discriminating explanations, even when objects partially overlap. Additionally, FGVis is able to visualize fine-grained details down to the pixel level.}
	\label{fig:fine_grained_class_discriminative}
	\vspace{\belowfigskip}
\end{figure}

Visual explanation methods should be able to produce class discriminative (\ie focus on one object) and fine-grained explanations~\cite{selvaraju2017grad}. To test FGVis with respect to these properties, we generate explanations for images containing two objects. The objects are chosen from highly different categories to ensure little overlapping evidence. In Fig.~\ref{fig:fine_grained_class_discriminative}, we visualize explanations of three such images, computed using the \textit{deletion game} and \textit{GoogleNet}. Additional results can be found in Sec.~\ref{sec:appx_classDiscriminative}.

FGVis is able to generate class discriminative explanations and only highlights pixels of the chosen target class. Even partially overlapping objects, as the elkhound and ball in Fig.~\ref{fig:fine_grained_class_discriminative}, first row, or the bridge and schooner in Fig.~\ref{fig:fine_grained_class_discriminative}, third row, are correctly discriminated. One major advantage of FGVis is its ability to visualize fine-grained details. This property is especially visible in Fig~\ref{fig:fine_grained_class_discriminative}, second row, which shows an explanation for the target class fence. Despite the fine structure of the fence, FGVis is able to compute a precise explanation which mainly contains fence pixels.

\begin{figure*}[t!]
	\centering
	\includegraphics[width=0.9\linewidth]{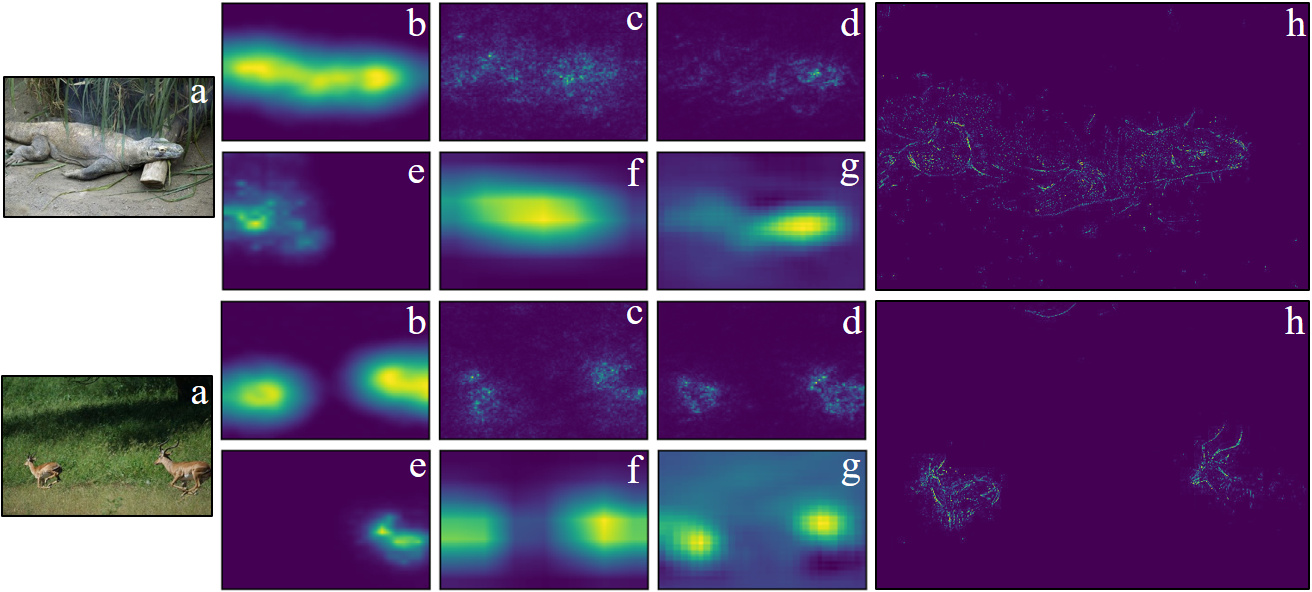}
	\caption{Comparison of mean explanation masks: a) Image, b) BBMP~\cite{fong2017interpretable}, c) Gradient~\cite{simonyan2013deep}, d) Guided Backprop~\cite{springenberg2015striving} , e) Contrastive Excitation Backprop~\cite{zhang2016top}, f) Grad-CAM~\cite{selvaraju2017grad}, g) Occlusion~\cite{zeiler2014visualizing}, h) FGVis (ours). The masks of all reference methods are based on work by~\cite{fong2017interpretable}. Due to our detailed and sparse masks, we plot them in a larger size.}
	\label{fig:comparison}
	\vspace{\belowfigskip}
\end{figure*}
\begin{figure}[t!]
	\centering
	\includegraphics[width=0.95\linewidth]{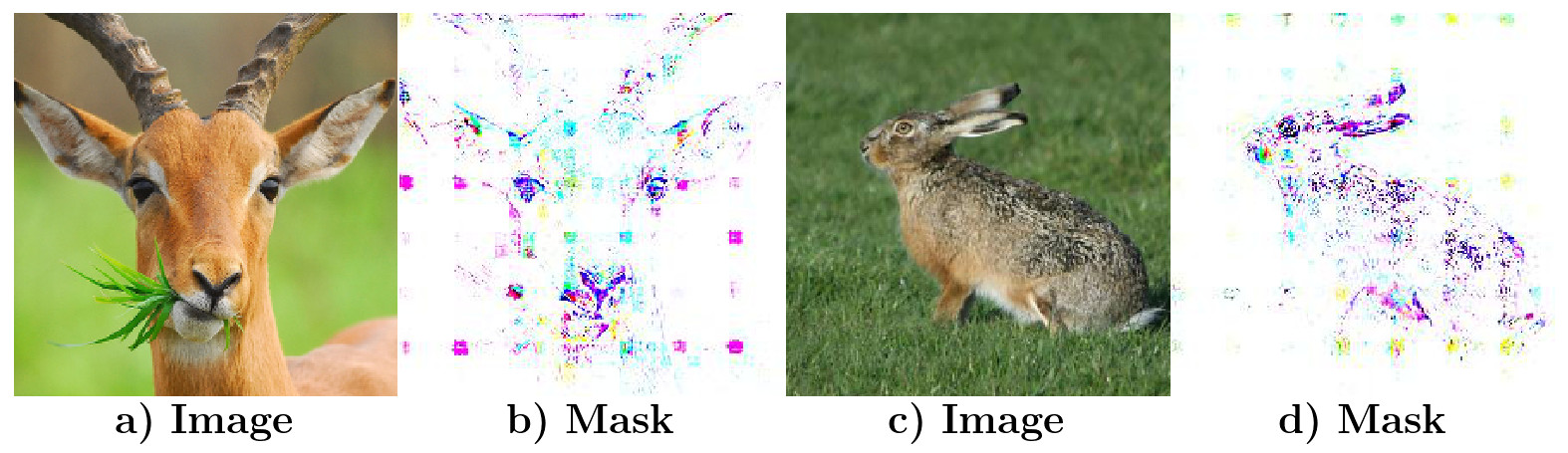}
	\caption{Visual explanations computed using the \textit{deletion game} for \textit{ResNet50}. The masks (b, d) show a grid-like pattern, as also observed in~\cite{nie2018theoretical} for \textit{ResNet50}.}
	\label{fig:resnet_grid_pattern}
	\vspace{\belowfigskip}
\end{figure}

%% file: bias_training_data.tex
\subsection{Investigating Biases of Training Data}
\label{sec:biases}
An application of explanation methods is to identify a bias in the training data. Especially for safety-critical, high-risk domains (\eg autonomous driving), such a bias can lead to failures if the model does not generalize to the real world.

\begin{figure}[t!]
	\centering
	\includegraphics[width=0.92\linewidth]{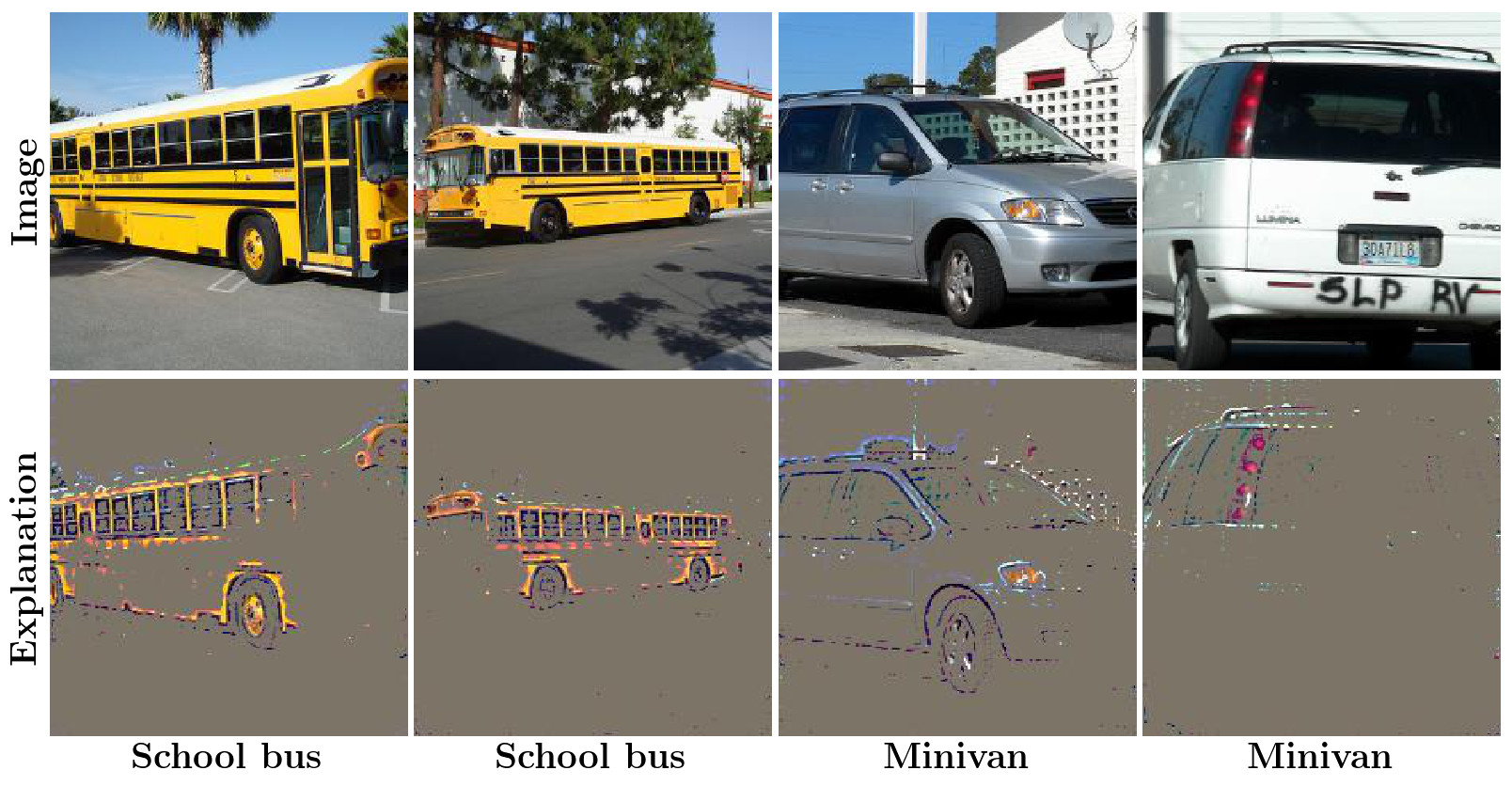}
	\vspace{-0.1cm}
	\caption{Explanations computed using the \textit{preservation game} for \textit{VGG16}. Explanations of the class minivan focus on edges, hardly preserving the color, compared to the class school bus, with yellow dominating the explanations.}
	\label{fig:color_bias}
	\vspace{\belowfigskip}
\end{figure}

\textbf{Learned objects.} One common bias is the coexistence of objects in images which can be depicted using FGVis. In Sec.~\ref{sec:appx_biases}, we describe such a bias in ImageNet for sports equipment appearing in combination with players. 
\\
\indent\textbf{Learned color.} Objects are often biased towards specific colors. FGVis can give a first visual indication for the importance of different color channels. We investigate if a \textit{VGG16} model trained on ImageNet shows such a bias using the \textit{preservation game}. We focus on images of school buses and minivans and compare explanations (Fig.~\ref{fig:color_bias}; all correctly predicted images in Fig.~\ref{fig_apx:schoolbus_expl_all} and~\ref{fig_apx:minivan_expl_all}). Explanations of minivans focus on edges, not consistently preserving the color compared to school buses with yellow dominating those explanations. This is a first indication for the importance of color for the prediction of school buses. 

To verify the qualitative finding, we quantitatively give an estimation of the color bias. As an evaluation we swap each of the three color channels \textit{BGR} to either \textit{RBG} or \textit{GRB} and calculate the ratio of maintained true classifications on the validation data after the swap. For minivans $83.3\%$ (averaged over \textit{RBG} and \textit{GRB}) of the $21$ correctly classified images keep their class label, for school buses it is only $8.3\%$ of $42$ images. For $80$ ImageNet classes at least $75\%$ of images are no longer truly classified after the color swap. We show the results for the most and least affected $19$ classes and minivan / school bus in Tab.~\ref{table:color_bias}. 
\\
To the best of our knowledge, FGVis is the first method used to highlight color channel importance. 

%% file: faithfulness_explanations.tex
\subsection{Faithfulness of Explanations}
\label{sec:faithfulness}
The faithfulness of generated visual explanations to the underlying neural network is an important property of explanation methods~\cite{selvaraju2017grad}. To quantitatively compare the faithfulness of methods, Petsiuk~\etal~\cite{petsiuk2018rise} proposed causal metrics which do not depend on human labels. These metrics are not biased towards human perception and are thus well suited to verify if an explanation correctly represents the evidence on which a model bases its prediction. 

We use the deletion metric~\cite{petsiuk2018rise} to evaluate the faithfulnes of explanations generated by our method. This metric measures how the removal of evidence effects the prediction of the used model. The metric assumes that an importance map is given, which ranks all image pixels with respect to their evidence for the predicted class $c_{ml}$. By iteratively removing important pixels from the input image and measuring the resulting probability of the class $c_{ml}$ a deletion curve can be generated, whose \textit{area under the curve} AUC is used as a measure of faithfulness (Sec.~\ref{sec:appx_faithfulnes}). 

In Tab.~\ref{table:faithfullness}, we report the deletion metric of FGVis, computed on the validation split of ImageNet using different models. We use the \textit{deletion game} to generate masks $\mathbf{m}_{ml}$, which determine the importance of each pixel. A detailed description of the experiment settings as well as additional figures, can be found in Sec.~\ref{sec:appx_faithfulnes}. FGVis outperforms the other explanation methods on both models by a large margin. This performance increase can be attributed to the ability of FGVis to visualize fine-grained evidence. All other approaches are limited to coarse explanations, either due to computational constraints or due to the used measures to avoid adversarial evidence. The difference between the two model architectures can most likely be attributed to the superior performance of \textit{ResNet50}, resulting in on average higher softmax scores over all validation images.
\begin{table}[h!]
	\centering
	\scalebox{0.9}{
		\begin{tabular}{ccc}
			\hline
			Method 				 							& \textit{ResNet50}  	& \textit{VGG16} \\
			\hline
			Grad-Cam~\cite{selvaraju2017grad} 				& $0.1232$ 			& $0.1087$ \\
			Sliding Window~\cite{zeiler2014visualizing} 	& $0.1421$ 			& $0.1158$ \\
			LIME\cite{ribeiro2016should}  				 	& $0.1217$ 			& $0.1014$ \\
			RISE~\cite{petsiuk2018rise} 				 	& $0.1076$ 			& $0.0980$ \\
			\hline
			FGVis (ours)		 							& $\mathbf{0.0644}$ 	& $\mathbf{0.0636}$ \\
			\hline
		\end{tabular}
	}
	\caption{Deletion metric computed on the ImageNet validation dataset (lower is better). The results for all reference methods were taken from Petsiuk~\etal~\cite{petsiuk2018rise}.}
	\label{table:faithfullness}
	\vspace{\belowfigskip}
\end{table}

%% file: retina.tex
\subsection{Visual explanation for medical images}
\label{sec:retina}
We evaluate FGVis on a real-world use case to identify regions in eye fundus images which lead a CNN to classify the image as being affected with referable diabetic retinopathy (RDR). Using the \textit{deletion game} we derive a weakly-supervised approach to detect RDR lesions. The setup, used network, as well as details on the disease and training data are described in~\ref{sec:retina_appx}. To evaluate FGVis, the DiaretDB1 dataset \cite{kauppi2007diaretdb1} is used containing 89 fundus images with different lesion types, ground truth marked by four experts. To quantitatively judge the performance, we compare in Tab.~\ref{tab:retina_img_level_se} the image level sensitivity of detecting if a certain lesion type is present in an image. The methods~\cite{zhou2016automatic, liu2017location, haloi2015gaussian, mane2015detection} use supervised approaches on image level without reporting a localization. \cite{zhao2018uniqueness} propose an unsupervised approach to extract salient regions. \cite{gondal2017weakly} use a comparable setting to ours applying CAM~\cite{zhou2016learning} in a weakly-supervised way to highlight important regions. To decide if a lesion is detected, \cite{gondal2017weakly} suggest an overlap of 50\% between proposed regions and ground truth. As our explanation masks are fine-grained and the ground truth is coarse, we compare using a 25\% overlap and for completeness report a 50\% overlap. 

It is remarkable that FGVis performs comparable or outperforms fully supervised approaches which are designed to detect the presence of one lesion type. The strength of FGVis is especially visible in detecting RSD, as these small lesions only cover some pixels in the image. In Fig.~\ref{fig:diaretdb1} we show fundus images, ground truth and our predictions. 

\begin{table} [h!]
	\centering
	\begin{threeparttable}
		\scalebox{0.9}{
			\begin{tabular}{c*{5}{>{}c<{}}}
				\hline
				Method & H & HE & SE & RSD  \\
				\hline
				Zhou \textit{et al.}\cite{zhou2016automatic}& 94.4 & - & - \\
				Liu \textit{et al.}\cite{liu2017location} &- & 83.0 & 83.0 & - \\
				Haloi \textit{et al.}\cite{haloi2015gaussian} & & \textbf{96.5} & - & - \\
				Mane \textit{et al.}\cite{mane2015detection}  & -& - & - & \textbf{96.4} \\
				Zhao \textit{et al.} \cite{zhao2018uniqueness}& 98.1 & - & - &\\
				Gondal \textit{et al.}\cite{gondal2017weakly} & 97.2 & 93.3 & 81.8 & 50 \\
				\hline
				Ours (25\% Overlap) & \textbf{100} & 94.7 & \textbf{90.0} & 88.4  \\ 
				Ours (50\% Overlap) & 90.5 & 81.6 & 80.0 & 86.0  \\ 
				\hline
			\end{tabular}
		}
	\end{threeparttable}
	\caption{Image level sensitivity in \% (higher is better) for four different lesions H, HE, SE, RSD: Hemorrhages, Hard Exudates, Soft Exudates and Red Small Dots.} 
	\label{tab:retina_img_level_se}
	\vspace{\belowfigskip}
\end{table}

%% file: conclusion.tex
\section{Conclusion}
We propose a method which generates fine-grained visual explanations in the image space using on a novel technique to defend adversarial evidence. Our defense does not introduce hyperparameters. We show the effectivity of the defense on different models, compare our explanations to other methods, and quantitatively evaluate the faithfulness. Moreover, we underline the strength in producing class discriminative visualizations and point to characteristics in explanations of a \textit{ResNet50}.  Due to the fine-grained nature of our explanations, we achieve remarkable results on a medical dataset. Besides, we show the usability of our approach to visually indicate a color bias in training~data. 
\clearpage
\newpage

%% file: appendix.tex
\title{Interpretable and Fine-Grained Visual Explanations for \\Convolutional Neural Networks\\-Supplementary Material-}

\author{J\"org Wagner$^{1, 2}$ \hspace{7mm} Jan Mathias K\"ohler$^{1}$ \hspace{7mm} Tobias Gindele$^{1,}$\footnotemark[1] \hspace{7mm} Leon Hetzel$^{1,}$\footnotemark[1] \\ Jakob Thadd\"aus Wiedemer$^{1,}$\footnotemark[1]\ \hspace{7mm} Sven Behnke$^{2}$\\
	$^1$Bosch Center for Artificial Intelligence (BCAI), Germany \hspace{6mm} $^2$University of Bonn, Germany\\
	{\tt\small Joerg.Wagner3@de.bosch.com; behnke@cs.uni-bonn.de}
}

\maketitle
\thispagestyle{empty}

\renewcommand{\thetable}{A\arabic{table}}
\renewcommand{\thefigure}{A\arabic{figure}}
\renewcommand{\thesection}{A\arabic{section}}

\setcounter{table}{0}
\setcounter{figure}{0}
\setcounter{section}{0}
\setcounter{page}{1} 
 
The supplementary material provides details, additional results, and further comparisons. 
\section{Defending against Adversarial Evidence}

Our method produces explanations based on evidence in the image and suppresses hallucination of adversarial evidence. Without our adversarial defense the optimization can produce an explanation for any class (\ie even for a class visually not present in the image). 

\begin{figure}[t!]
	\begin{center}
		\includegraphics[width=0.8\linewidth]{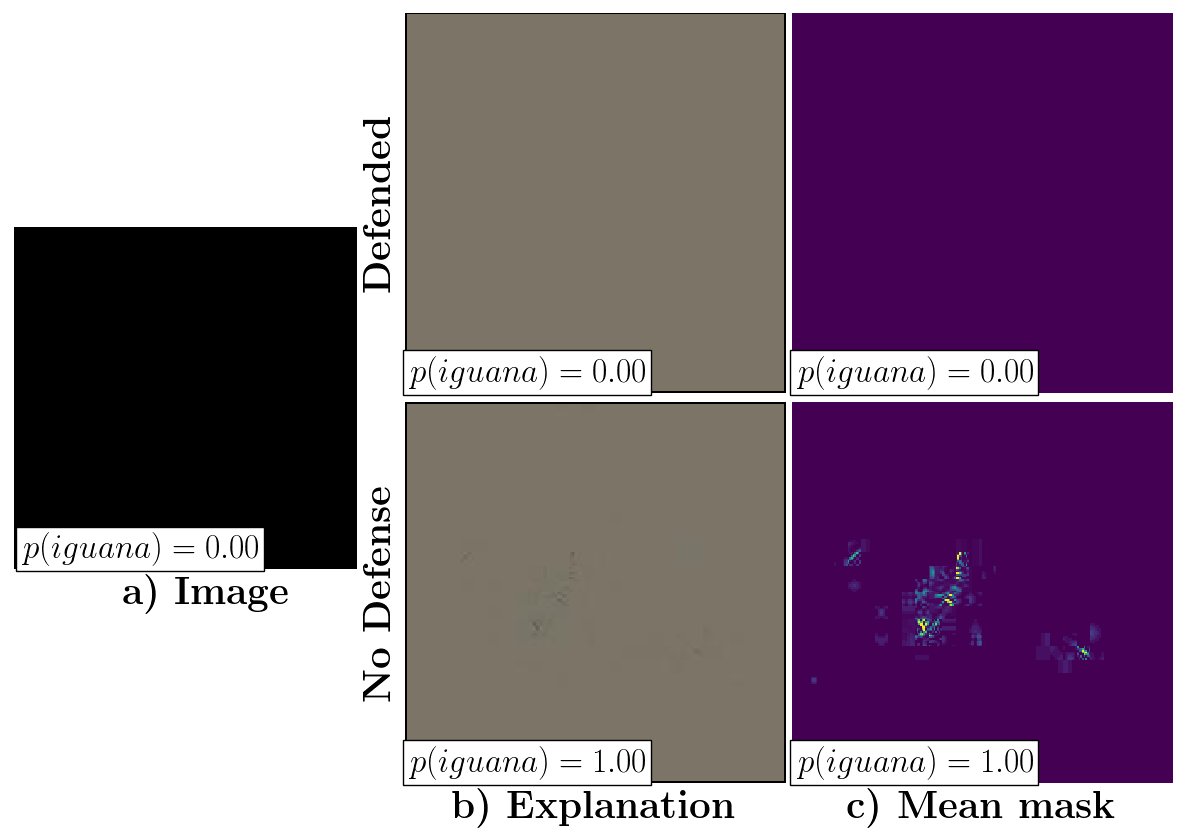}
	\end{center}
	\vspace{-0.5cm}
	\caption{Explanation for the adversarial class \textit{iguana} starting from a black image. An adversarial can only be computed without defense (\textit{generation game}, \textit{GoogleNet}). Mean masks are enhanced by a factor of 10.}
	\label{fig:appx_adversarial_evidence}
\end{figure}

To illustrate this differently to the experiment reported in Sec.~\ref{sec:perturbation_based_explanations} (Tab.~\ref{table:adversarial_defense} and Fig.~\ref{fig:adversarial_evidence}), we show an alternative version of the evaluation, only using a black image as input. Fig.~\ref{fig:appx_adversarial_evidence} shows an explanation for the adversarial class \textit{iguana} with and without defense. For Tab.~\ref{table:appx_adversarial_defense} we create explanations for each of the 998 ImageNet classes, using always the same black input image.  We omit the predicted class of the black image and the class of the starting condition (image $\cdot$ zero mask). Without defense an explanation can always be generated due to hallucination of adversarial evidence. The results are comparable to the evaluation in the main paper. 

\section{Implementation Details} \label{sec:appx_implementation}
Unless otherwise specified, the explanations are computed for the most-likely class using SGD with a learning rate of $0.1$, running for $500$ iterations. To improve optimization and avoid instabilities, we initialize the masks $\mathbf{m}$ with noise sampled for each pixel from a uniform distribution $\mathcal{U}(a,b)$.  with $\mathcal{U}(0, 0.01)$ for the \textit{generation} and \textit{repression game} and $\mathcal{U}(0.99, 1)$ for the \textit{preservation} and \textit{deletion game}. We normalize the gradient using its maximum value to avoid large changes of individual mask pixels. 

\begin{table}[t!]
	\centering
	\scalebox{0.9}{
		\begin{tabular}{ccccc}
			\hline
			Model 		& \textit{GoogleNet}	& \textit{VGG16} 	& \textit{AlexNet} 	& \textit{ResNet50}  \\
			\hline 
			No Defense 	& $100\,\%$ 			& $100\,\%$			& $100\,\%$			& $100\,\%$			 \\
			Defended 	& $0.0\,\%$ 			& $0.1\,\%$			& $0.1\,\%$		& $0.0\,\%$		 \\
			\hline
		\end{tabular}
	}
	\caption{How often an adversarial class could be generated from a black image averaged over 998 ImageNet classes (\textit{generation game}, $\lambda = 0$).}
	\label{table:appx_adversarial_defense}
	\vspace{-0.2cm}
\end{table}

For the similarity metric $\varphi(\cdot, \cdot)$ we use the cross-entropy for the \textit{generation} and \textit{preservation game} and the negative probability for the \textit{deletion} and \textit{repression game}. 

When computing an explanation for the most-likely class, we use a line-search for the parameter $\lambda$ to determine its optimal value. Unless otherwise noted, we iteratively use $13$ equally spaced $\lambda$ values between $10^{-4}$ and $10^{-10}$ and stop when the resulting most-likely class of $\mathbf{e}_{ml}$ shifts (\textit{deletion} and \textit{repression game}) or achieves the highest probability among all classes (\textit{preservation} and \textit{generation game}). We use images of the ImageNet~\cite{krizhevsky2012imagenet} validation set and pre-trained model weights.

\begin{figure*}[b!]
	\centering
	\begin{subfigure}{.15\linewidth}
		\centering
		\includegraphics[width=1.0\linewidth]{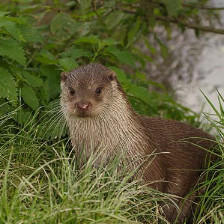}
		\caption{Image}
	\end{subfigure}%
	\begin{subfigure}{.82\linewidth}
		\centering
		\includegraphics[trim=5cm 4.5cm 5cm 4cm, clip,width=1.0\linewidth]{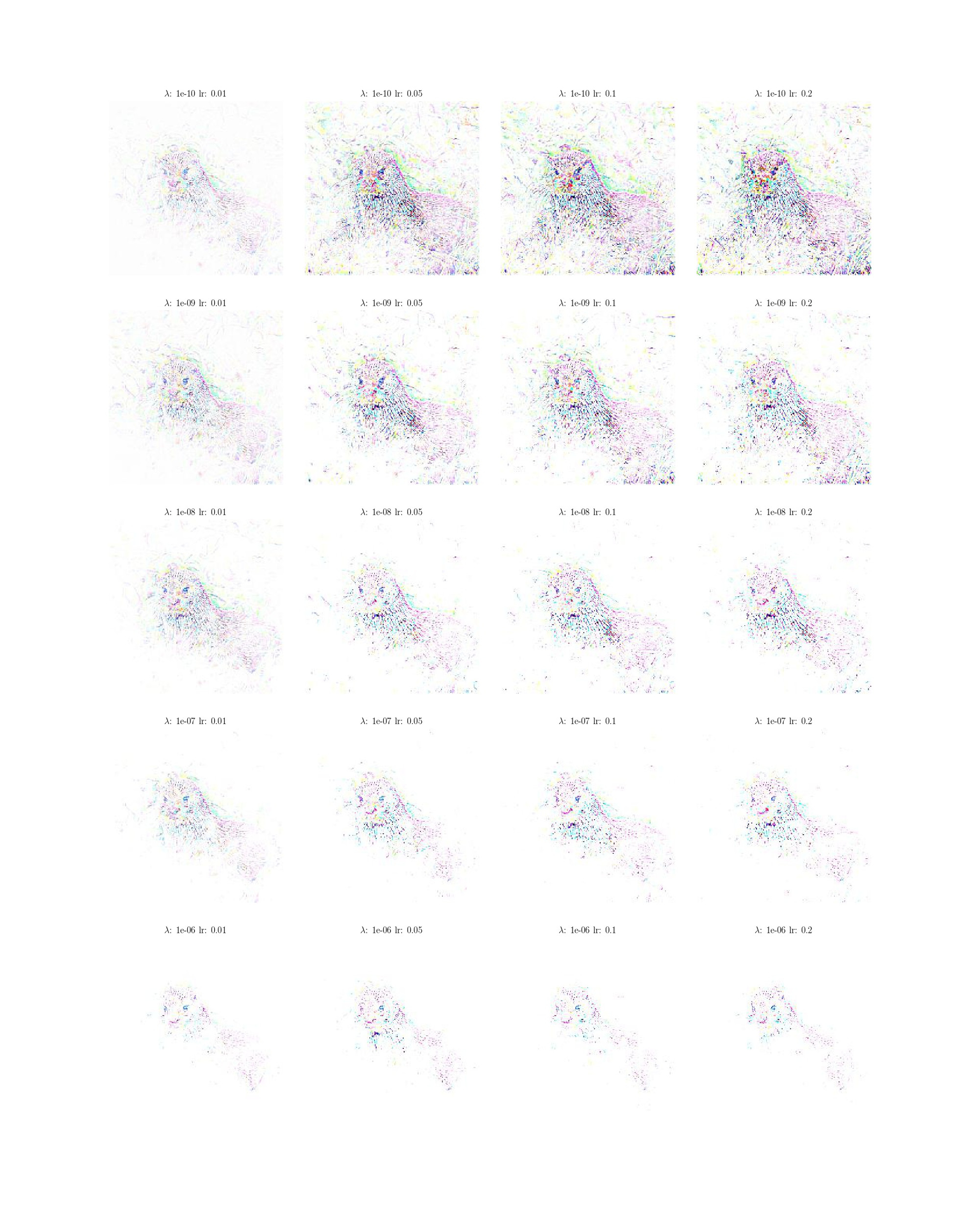}
		\caption{Masks of class \textit{otter}.}
	\end{subfigure}
	\caption{Comparison of resulting masks for different learning rates (lr) and $\lambda$ values computed using the \textit{deletion game} and \textit{GoogleNet}.}
	\label{fig_apx:hyperparameter}
\end{figure*}

A comparison of resulting masks for different learning rates and $\lambda$ values for \textit{GoogleNet} computed with the \textit{deletion game} are shown in~Fig.~\ref{fig_apx:hyperparameter}. 

A higher $\lambda$ value causes sparser masks due to a higher weighting of the sparsity invoking part $\left\| \mathbf{m}_{C_T} \right\|_1$ within the loss function (Eq.~\ref{eq:preservation_game} and Eq.~\ref{eq:deletion_game}). Especially for higher $\lambda$ values, the resulting masks are rather independent of the chosen learning rate of the SGD optimization.

\section{Qualitative Results} \label{sec:appx_qualitative_results}
\subsection{Entropy of Reference Images} \label{sec:appx_references}
FGVis computes explanations $\mathbf{e}_{c_T}$ by optimizing for a perturbed version of the input image $\mathbf{x}$. The perturbation is modelled via a removal operator $\Phi$~\cite{fong2017interpretable, du2018towards, chang2018explaining, dabkowski2017real}, which computes a weighted average between the image $\mathbf{x}$ and a reference image $\mathbf{r}$, using a mask $\mathbf{m}_{C_T}$:
\begin{equation}
\mathbf{e}_{c_T} = \Phi(\mathbf{x}, \mathbf{m}_{c_T}) = \mathbf{x} \cdot \mathbf{m}_{c_T} + (1 - \mathbf{m}_{c_T}) \cdot \mathbf{r}.
\end{equation}
A good reference image $\mathbf{r}$ should carry little information and lead to a model prediction with a high entropy, meaning, ideally all classes are assigned the same softmax score (see '\textit{Maximum	(1000 classes)}' in Tab.~\ref{table:entropy} for the resulting maximum entropy). To compare references, we report their entropy for different models in Tab.~\ref{table:entropy}. 

For all models except \textit{GoogleNet} the zero image reference has the highest entropy. Interestingly, for the zero image reference, the more recent architectures (\textit{GoogleNet}, \textit{ResNet50}) have a lower entropy. This indicates that these architectures do not assign a roughly equally distributed softmax score to all classes (as \textit{AlexNet} or \textit{VGG16}).

As expected, an increasing noise level $\sigma_n$ for a Gaussian noise image as well as a decreasing standard deviation of the Gaussian blur filter $\sigma_b$ reduces the entropy. Only \textit{GoogleNet} does not fully follow this characteristic. 

For comparison, we report the entropy for 1000 random ImageNet validation images for the different models. 

Due to the high entropy as well as the low computational effort of a zero reference image, we choose this reference for FGVis. 

\begin{table*}[b!]
	\centering
	\scalebox{0.95}{
		\begin{tabular}{ccccc}
			\hline
			Reference image $\mathbf{r}$					& \textit{AlexNet} 	& \textit{GoogleNet} 	& \textit{VGG16} 	& \textit{ResNet50}	\\
			\hline	
			Zero image						 				& $6.90$ 			& $4.08$ 				& $6.31$ 			& $5.09$ 			\\
			Gaussian noise image ($\sigma_{n}=8$)			& $5.11 \pm 0.16$ 	& $4.62 \pm 0.16$ 		& $5.59 \pm 0.09$ 	& $4.56 \pm 0.14$ 	\\
			Gaussian noise image ($\sigma_{n}=32$)			& $2.61 \pm 0.29$ 	& $4.67 \pm 0.22$ 		& $4.38 \pm 0.23$ 	& $4.07 \pm 0.30$ 	\\
			Blurred ImageNet image ($\sigma_{b}=5$)			& $3.67 \pm 1.12$ 	& $3.15 \pm 1.31$ 		& $4.08 \pm 1.43$  	& $2.38 \pm 1.58$	\\
			Blurred ImageNet image ($\sigma_{b}=10$)		& $4.56 \pm 0.88$ 	& $4.09 \pm 1.08$ 		& $4.83 \pm 0.86$  	& $3.22 \pm 1.25$	\\
			\hline
			ImageNet image						 			& $1.73 \pm 1.43$ 	& $1.09 \pm 1.14$ 		& $1.06 \pm 1.22$	& $0.67 \pm 0.91$	\\
			Maximum	(1000 classes)							& $6.91$ 			& $6.91$ 				& $6.91$			& $6.91$			\\
			\hline
		\end{tabular}
	}	
	\caption{Entropy of reference images $\mathbf{r}$ for different models. The entropy is averaged over $1000$ random instances of each reference image. Gaussian noise images are generated by independently sampling for each pixel from a Gaussian distribution with zero-mean and a standard deviation of $\sigma_{n}$. The blurred ImageNet images are computed using a Gaussian blur filter with a standard deviation of $\sigma_{b}$. For all random references we report the mean $\pm$ standard deviation of the entropy.}
	\label{table:entropy}
\end{table*}

\subsection{Class Discriminative / Fine-Grained} \label{sec:appx_classDiscriminative}
\begin{figure*}[h!]
	\centering
	\includegraphics[width=0.85\linewidth]{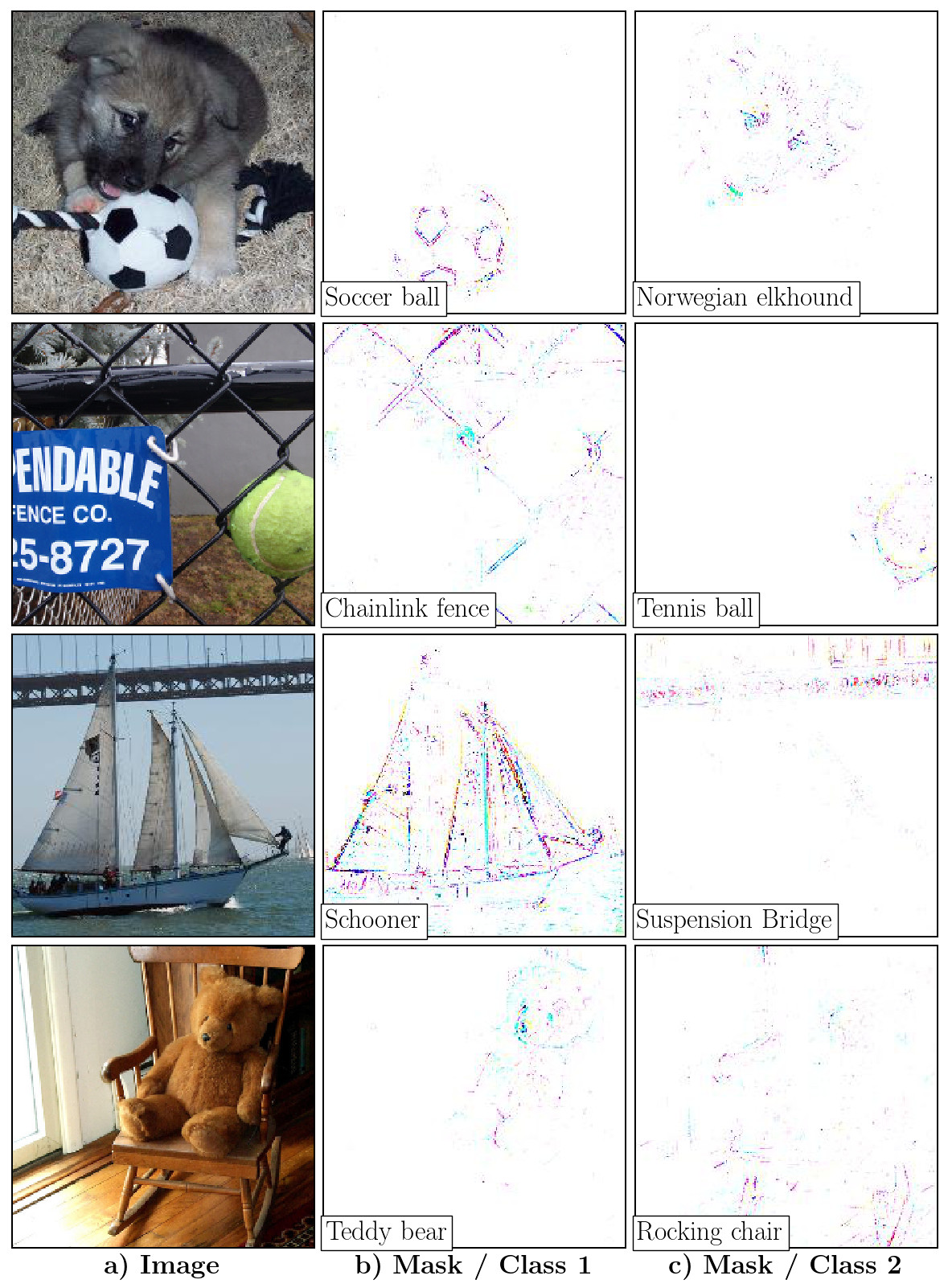}
	\caption{Explanation masks for images with multiple objects computed using the \textit{deletion game} and \textit{GoogleNet}. FGVis produces class discriminative explanations, even when objects partially overlap. Note that objects not belonging to either class, \eg the rug in the top row, the blue sign on the chainlink fence, or the window in the bottom row vanish in the explanation. Additionally, FGVis is able to visualize fine-grained details down to the pixel level. }
	\label{fig_apx:2class_1}
\end{figure*}

\begin{figure*}[h!]
	\centering
	\includegraphics[width=0.85\linewidth]{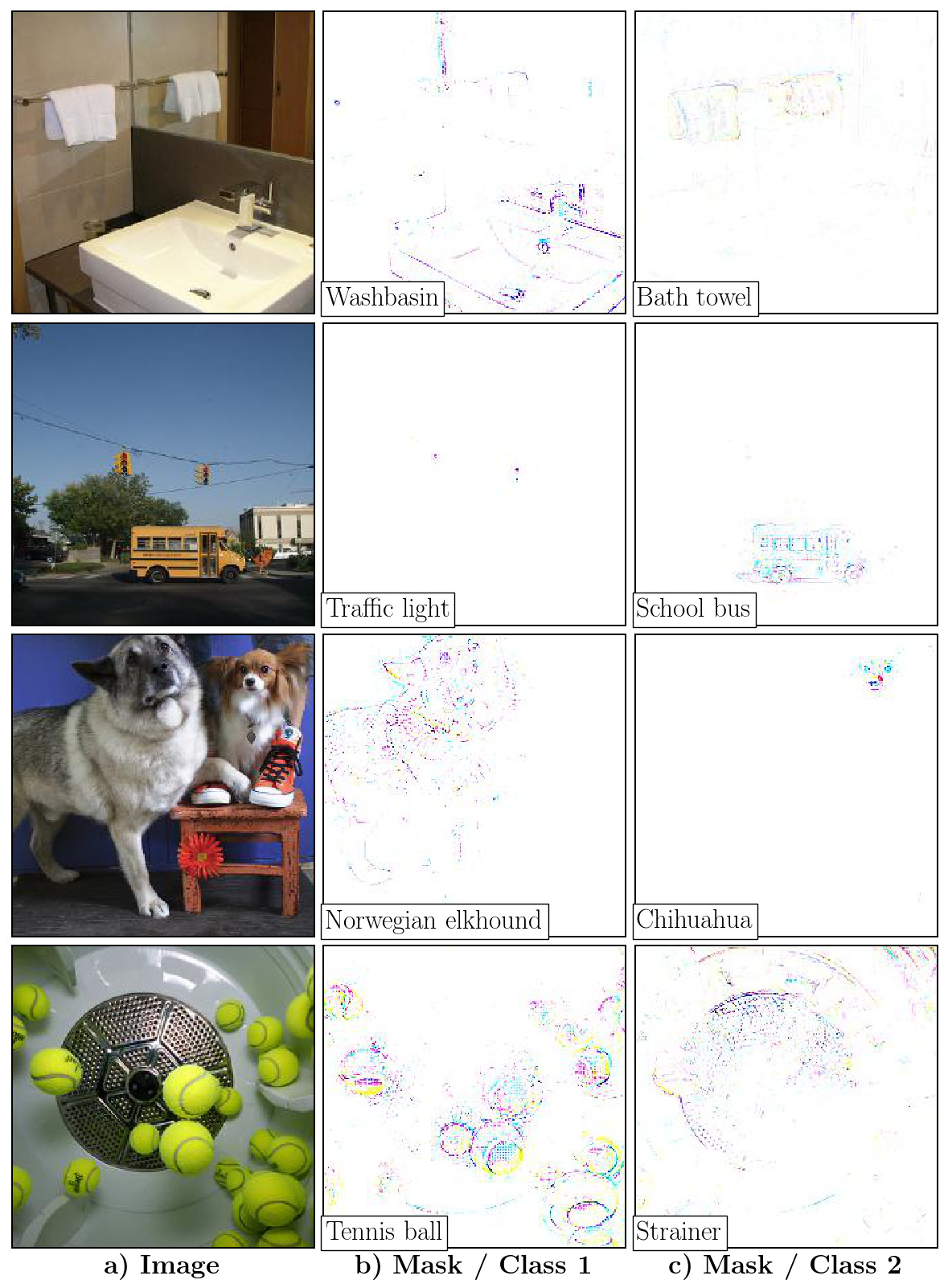}
	\caption{Explanation masks for images with multiple objects computed using the \textit{deletion game} and \textit{GoogleNet}. FGVis produces class discriminative explanations, even when objects partially overlap. This is especially visible in the last row where the tennis balls are almost all removed in the explanation mask for the class strainer.}
	\label{fig_apx:2class_2}
\end{figure*}

In Fig.~\ref{fig_apx:2class_1} and Fig.~\ref{fig_apx:2class_2}  we show additional explanation masks for images containing two distinct objects. The objects are chosen from highly different categories to ensure little overlapping evidence. The explanations are computed using the \textit{deletion game}, which generates the most pleasing class-discriminative explanations, and \textit{GoogleNet}. 

Note that FGVis discriminates well even if the two objects partially overlap. The figures additionally highlight the ability of FGVis to generate fine-grained explanations. 

To determine $\lambda$ we use for the most-likely class the strategy as described in Sec.~\ref{sec:appx_implementation}. For the second class $\lambda$ is optimized to significantly drop the softmax score of this class. 

\subsection{Investigating Biases of Training Data}\label{sec:appx_biases}
\begin{figure*}[t!]
	\vspace{1cm}
	\includegraphics[width=\linewidth]{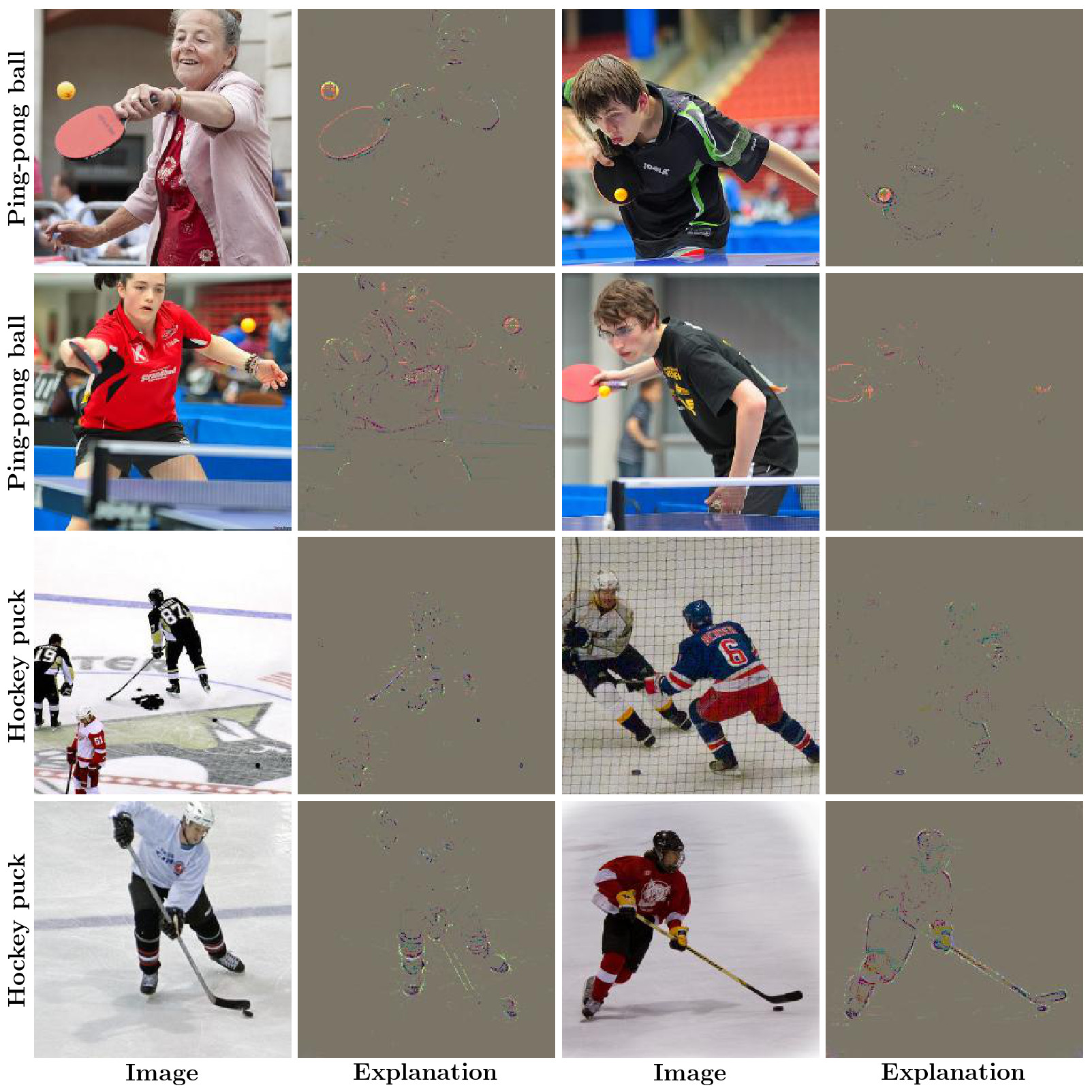}
	\caption{Visual explanations computed using the \textit{deletion game} for \textit{GoogleNet}. For both classes (hockey puck and ping-pong ball) the explanation method has to additionally delete pixels of the players and the table tennis bat/ice-hockey stick to shift the prediction of the model. This clearly highlights a bias of the data towards images which contain a puck/ball, a player and sports equipment.}
	\label{fig_apx:data_bias}
	\vspace{1cm}
\end{figure*}

\noindent\textbf{Learned objects.}
The coexistence of objects in images often results in a learned bias. In Fig.~\ref{fig_apx:data_bias}, we visualize such a bias for  \textit{GoogleNet} trained on ImageNet. 

Sports equipment like hockey pucks or ping-pong balls frequently appear in combination with players. This bias is learned by the neural network and results in explanations that also contain pixels belonging to the players. Without deleting these pixels, the \textit{deletion game} is not able to shift the class of the images.

\noindent\textbf{Learned color.}
We quantitatively verify the color bias reported in~Sec.~\ref{sec:biases} and show the $19$ classes of ImageNet which are most and least affected by swapping the color in Tab.~\ref{table:color_bias}. We swap each of the three color channels \textit{BGR} to either \textit{RBG} or \textit{GRB} and calculate the ratio of maintained true classifications on the validation data after the swap.

\begin{figure*}[h!]
	\centering
	\includegraphics[width=0.65\linewidth]{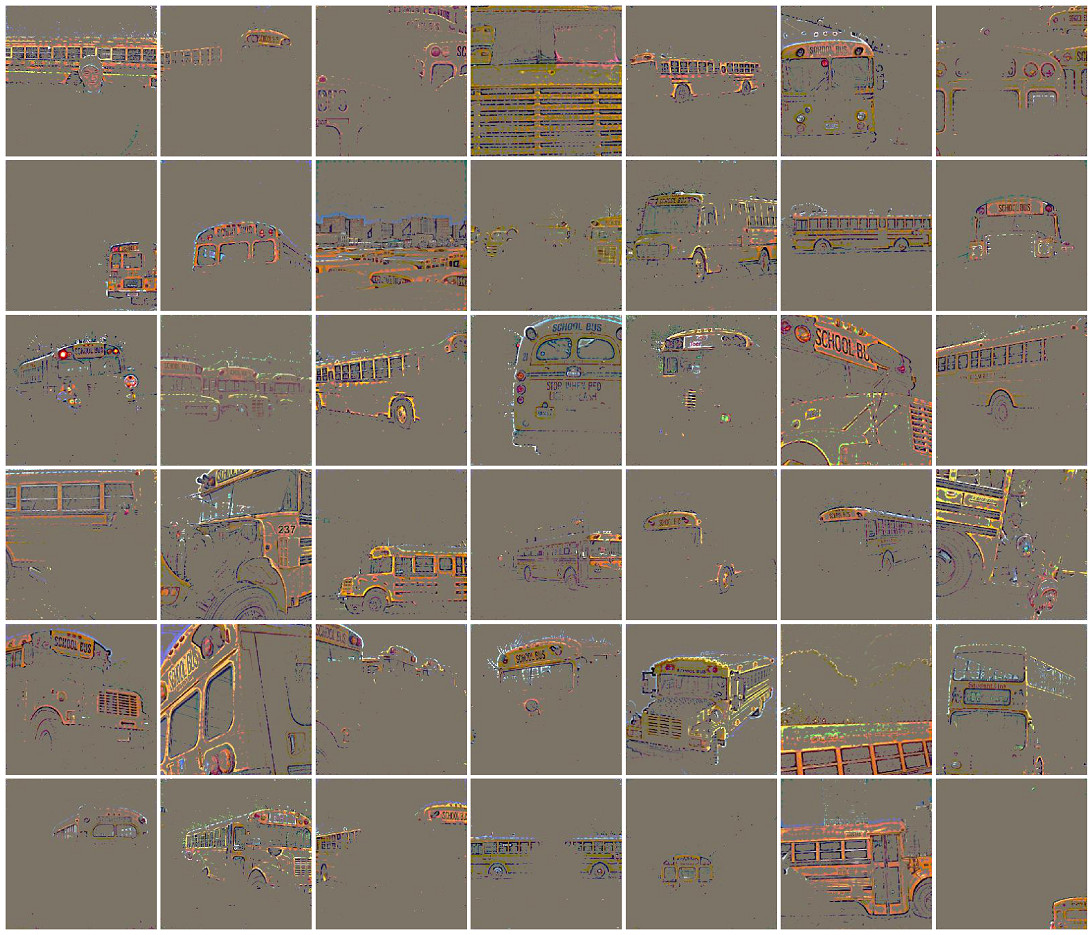}
	\caption{Explanations computed using the \textit{preservation game} for \textit{VGG} for the class school bus.}
	\label{fig_apx:schoolbus_expl_all}
	\vspace{0.5cm}
\end{figure*}

\begin{figure*}[h!]
	\centering
	\includegraphics[width=0.65\linewidth]{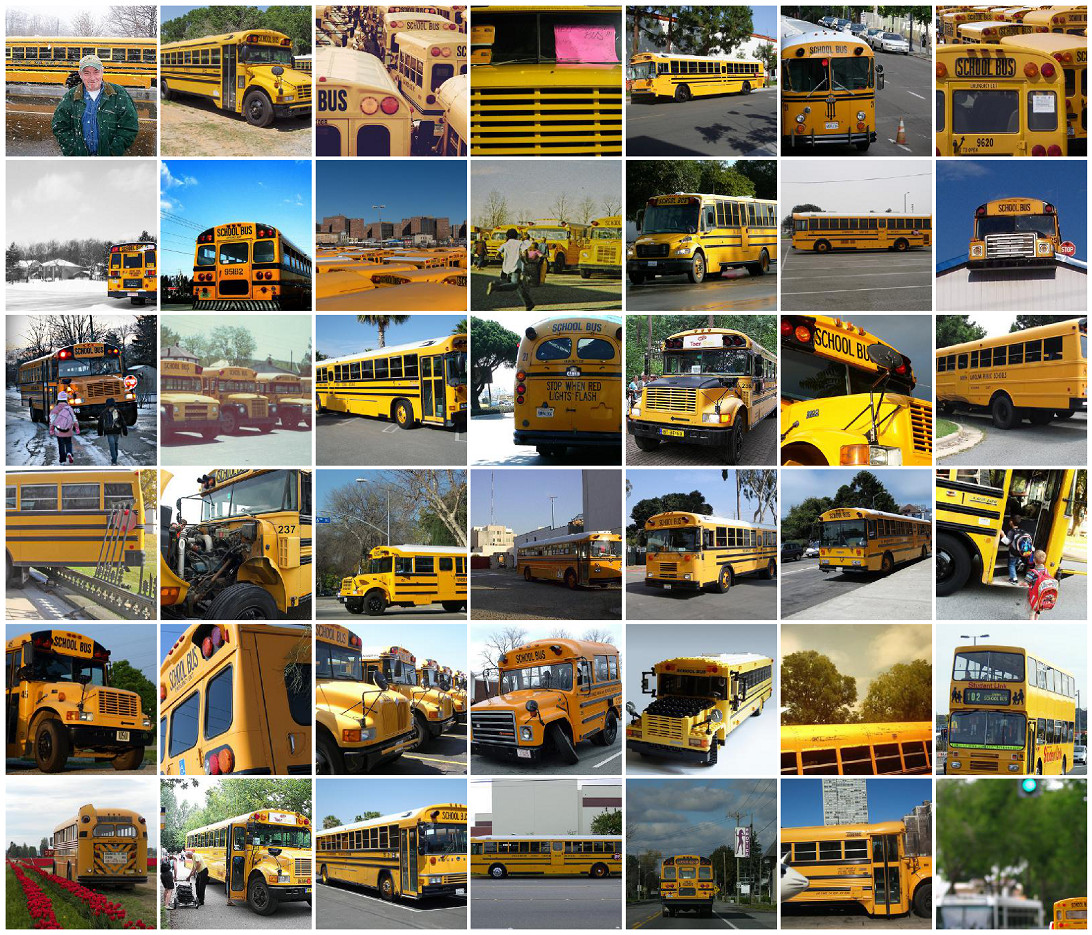}
	\caption{Input images for the explanations in Fig.~\ref{fig_apx:schoolbus_expl_all}}
	\label{fig_apx:schoolbus_imgs_all}
\end{figure*}

Fig.~\ref{fig_apx:schoolbus_expl_all} shows explanations for the class school bus computed using the \textit{preservation game} for \textit{VGG}. The yellow color, also visible in the original images (Fig.~\ref{fig_apx:schoolbus_imgs_all}), is dominant in most of the explanations. 

\begin{figure*}[h!]
	\centering
	\includegraphics[width=0.7\linewidth]{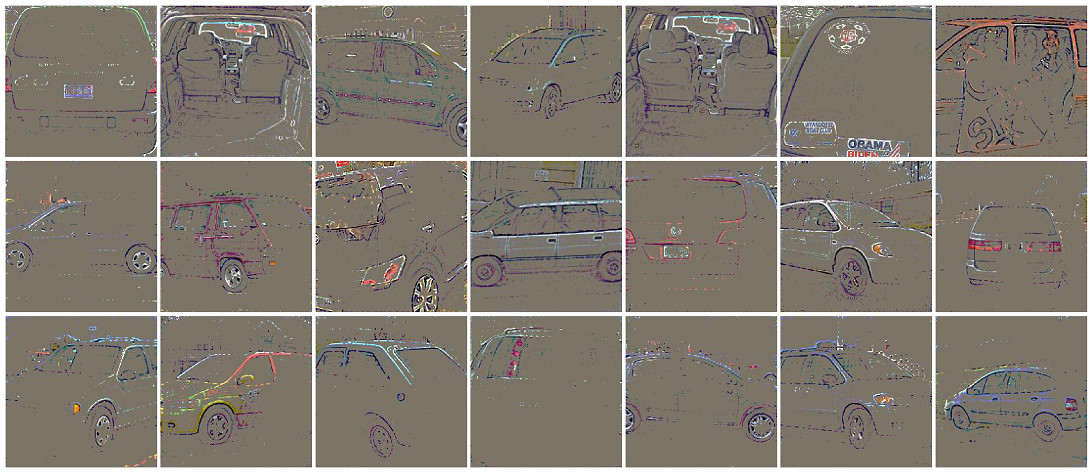}
	\caption{Explanations computed using the \textit{preservation game} for \textit{VGG} for the class minivan. }
	\label{fig_apx:minivan_expl_all}
	\vspace{0.7cm}
\end{figure*}

\begin{figure*}[h!]
	\centering
	\includegraphics[width=0.7\linewidth]{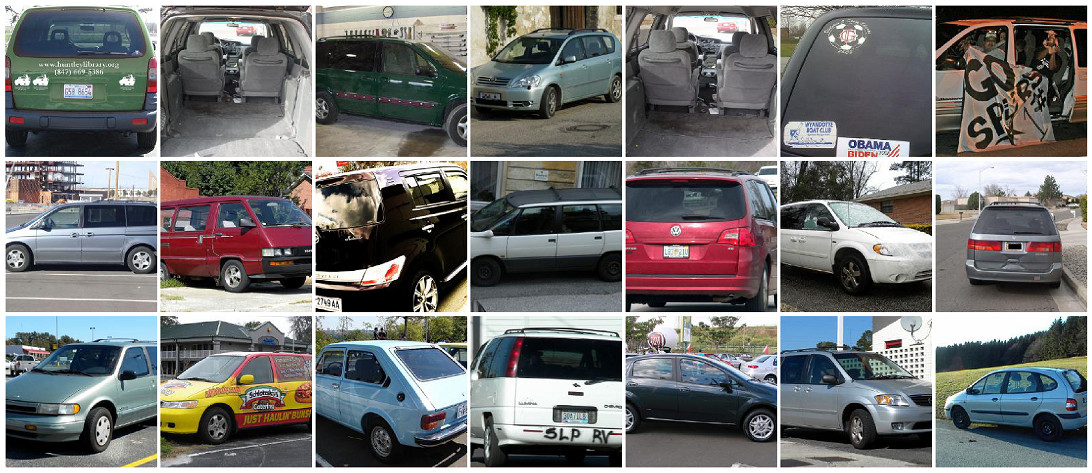}
	\caption{Input images for the explanations in Fig.~\ref{fig_apx:minivan_expl_all}}
	\label{fig_apx:minivan_imgs_all}
\end{figure*}

Fig.~\ref{fig_apx:minivan_expl_all} shows explanations for the class minivan computed using the \textit{preservation game} for \textit{VGG}. The original color of the car is not consistently preserved. Especially for white or grey cars (original images in Fig.~\ref{fig_apx:minivan_imgs_all}) the visible color in the explanation is reduced to a greenish-blue color.

Fig.~\ref{fig_apx:schoolbus_expl_all} and~\ref{fig_apx:minivan_expl_all} show all correctly classified images for school bus and minivan.  

\begin{table*}[hb!]
	\vspace{0.5cm}
	\centering
	\scalebox{1.0}{
		\begin{tabular}{rlrrrr}
			\hline
			ID &                               Class name &  \#Images &  		Avg. RBG, GRB &    	     RBG &   	    GRB \\
			\hline
			168 &                                  redbone &       31 &           $0.00\,\%$ &   $0.00\,\%$ &   $0.00\,\%$ \\
			964 &                                   potpie &       28 &           $0.00\,\%$ &   $0.00\,\%$ &   $0.00\,\%$ \\
			159 &                      Rhodesian ridgeback &       35 &           $0.00\,\%$ &   $0.00\,\%$ &   $0.00\,\%$ \\
			930 &                              French loaf &       27 &           $0.00\,\%$ &   $0.00\,\%$ &   $0.00\,\%$ \\
			234 &                               Rottweiler &       42 &           $1.19\,\%$ &   $0.00\,\%$ &   $2.38\,\%$ \\
			214 &                            Gordon setter &       36 &           $1.39\,\%$ &   $2.78\,\%$ &   $0.00\,\%$ \\
			963 &                         pizza, pizza pie &       35 &           $1.43\,\%$ &   $2.86\,\%$ &   $0.00\,\%$ \\
			950 &                                   orange &       35 &           $1.43\,\%$ &   $2.86\,\%$ &   $0.00\,\%$ \\
			184 &                            Irish terrier &       33 &           $1.52\,\%$ &   $0.00\,\%$ &   $3.03\,\%$ \\
			962 &                      meat loaf, meatloaf &       29 &           $1.72\,\%$ &   $3.45\,\%$ &   $0.00\,\%$ \\
			984 &                                 rapeseed &       47 &           $2.13\,\%$ &   $4.26\,\%$ &   $0.00\,\%$ \\
			211 &                vizsla, Hungarian pointer &       35 &           $2.86\,\%$ &   $2.86\,\%$ &   $2.86\,\%$ \\
			11 &           goldfinch, Carduelis carduelis &       48 &           $3.12\,\%$ &   $0.00\,\%$ &   $6.25\,\%$ \\
			934 &                 hotdog, hot dog, red hot &       40 &           $3.75\,\%$ &   $2.50\,\%$ &   $5.00\,\%$ \\
			218 &                   Welsh springer spaniel &       39 &           $3.85\,\%$ &   $2.56\,\%$ &   $5.13\,\%$ \\
			191 &               Airedale, Airedale terrier &       37 &           $5.41\,\%$ &   $5.41\,\%$ &   $5.41\,\%$ \\
			163 &                  bloodhound, sleuthhound &       18 &           $5.56\,\%$ &   $5.56\,\%$ &   $5.56\,\%$ \\
			961 &                                    dough &       15 &           $6.67\,\%$ &   $0.00\,\%$ &  $13.33\,\%$ \\
			263 &           Pembroke, Pembroke Welsh corgi &       41 &           $7.32\,\%$ &   $7.32\,\%$ &   $7.32\,\%$ \\
			\hline
			$\cdots$ & 						   $\cdots$ & $\cdots$ & 			 $\cdots$ & 	$\cdots$ & 	   $\cdots$ \\
			779 &                               school bus &       42 &           $8.33\,\%$ &   $9.52\,\%$ &   $7.14\,\%$ \\
			$\cdots$ & 						   $\cdots$ & $\cdots$ & 			 $\cdots$ & 	$\cdots$ & 	   $\cdots$ \\
			656 &                                  minivan &       21 &          $83.33\,\%$ &  $71.43\,\%$ &  $95.24\,\%$ \\
			$\cdots$ & 						   $\cdots$ & $\cdots$ & 			 $\cdots$ & 	$\cdots$ & 	   $\cdots$ \\
			\hline
			528 &               dial telephone, dial phone &       36 &          $95.83\,\%$ &  $91.67\,\%$ & $100.00\,\%$ \\
			866 &                                  tractor &       37 &          $95.95\,\%$ &  $91.89\,\%$ & $100.00\,\%$ \\
			572 &                                   goblet &       26 &          $96.15\,\%$ &  $96.15\,\%$ &  $96.15\,\%$ \\
			47 &  African chameleon, Chamaeleo chamaeleon &       40 &          $96.25\,\%$ &  $95.00\,\%$ &  $97.50\,\%$ \\
			302 &            ground beetle, carabid beetle &       27 &          $96.30\,\%$ &  $96.30\,\%$ &  $96.30\,\%$ \\
			463 &                             bucket, pail &       27 &          $96.30\,\%$ &  $96.30\,\%$ &  $96.30\,\%$ \\
			717 &                     pickup, pickup truck &       28 &          $96.43\,\%$ & $100.00\,\%$ &  $92.86\,\%$ \\
			178 &                               Weimaraner &       44 &          $96.59\,\%$ &  $93.18\,\%$ & $100.00\,\%$ \\
			669 &                             mosquito net &       44 &          $96.59\,\%$ &  $97.73\,\%$ &  $95.45\,\%$ \\
			661 &                                  Model T &       46 &          $96.74\,\%$ &  $97.83\,\%$ &  $95.65\,\%$ \\
			769 &                              rule, ruler &       36 &          $97.22\,\%$ & $100.00\,\%$ &  $94.44\,\%$ \\
			771 &                                     safe &       40 &          $97.50\,\%$ &  $97.50\,\%$ &  $97.50\,\%$ \\
			829 &   streetcar, tram, tramcar, trolley, ... &       41 &          $97.56\,\%$ &  $97.56\,\%$ &  $97.56\,\%$ \\
			713 &                              photocopier &       44 &          $97.73\,\%$ & $100.00\,\%$ &  $95.45\,\%$ \\
			916 &   web site, website, internet site, site &       47 &          $97.87\,\%$ & $100.00\,\%$ &  $95.74\,\%$ \\
			423 &                             barber chair &       31 &          $98.39\,\%$ &  $96.77\,\%$ & $100.00\,\%$ \\
			190 &               Sealyham terrier, Sealyham &       39 &          $98.72\,\%$ &  $97.44\,\%$ & $100.00\,\%$ \\
			340 &                                    zebra &       47 &         $100.00\,\%$ & $100.00\,\%$ & $100.00\,\%$ \\
			545 &                     electric fan, blower &       37 &         $100.00\,\%$ & $100.00\,\%$ & $100.00\,\%$ \\
			\hline
		\end{tabular}
	}
	\caption{Ratio of maintained true classifications on the validation data of ImageNet after swapping color channels for the most and least affected $19$ classes and minivan / school bus. Each of the three color channels \textit{BGR} are swapped to either \textit{RBG} or \textit{GRB}. The class ID, class name, number of truly classified images before the color swap (\#Images) and percentage of maintained classification after the swap for the average over \textit{RBG} or \textit{GRB} and each swap individually are reported. Most color-dependent classes are redbone or potpie. Most color-independent classes zebra or electric fan. }
	\label{table:color_bias}
	\vspace{0.5cm}
\end{table*}



\newpage

\subsection{Comparison of Networks} 
\label{sec:appx_comparisonNets}
\begin{figure*}[h!]
	\centering
	\includegraphics[width=0.95\linewidth]{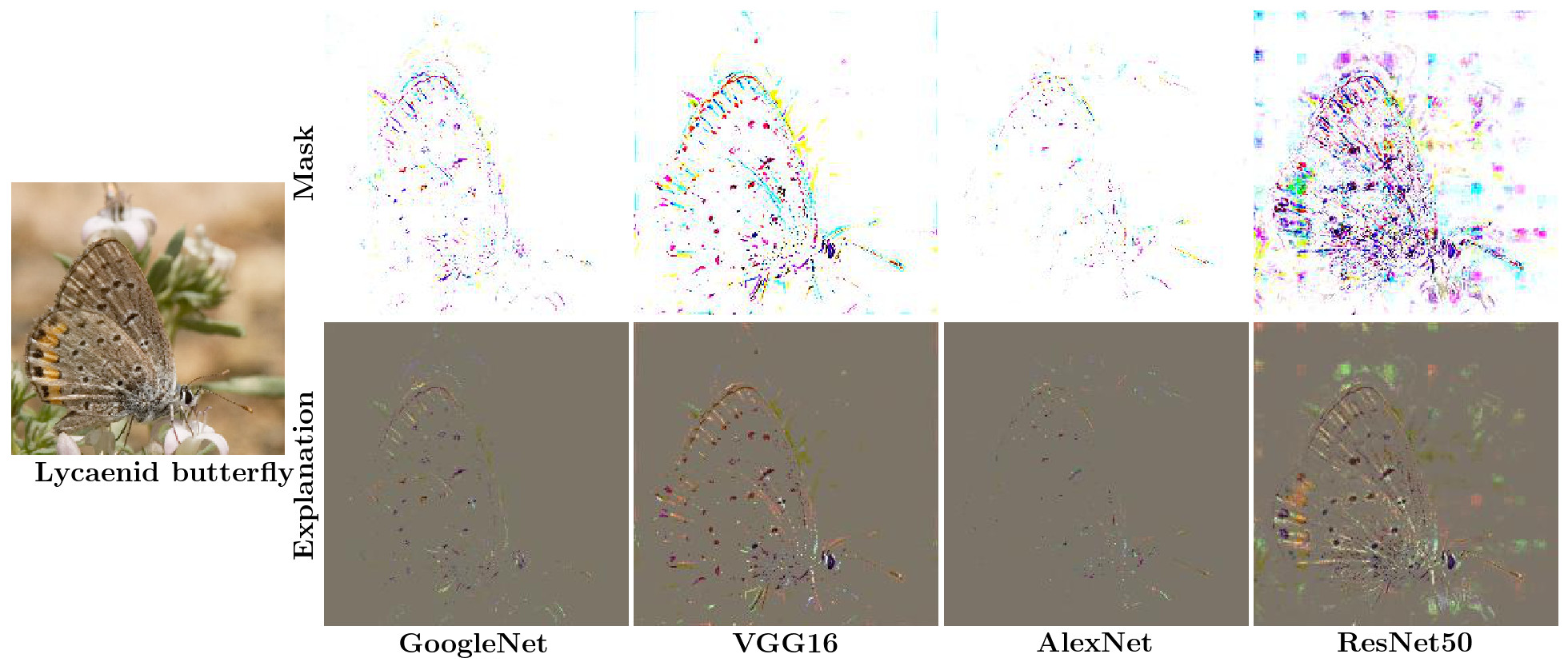}
	\caption{Masks and explanations computed using the \textit{deletion game} for different networks. }
	\label{fig_apx:lycaenid}
\end{figure*}

\begin{figure*}[h!]
	\centering
	\includegraphics[width=0.95\linewidth]{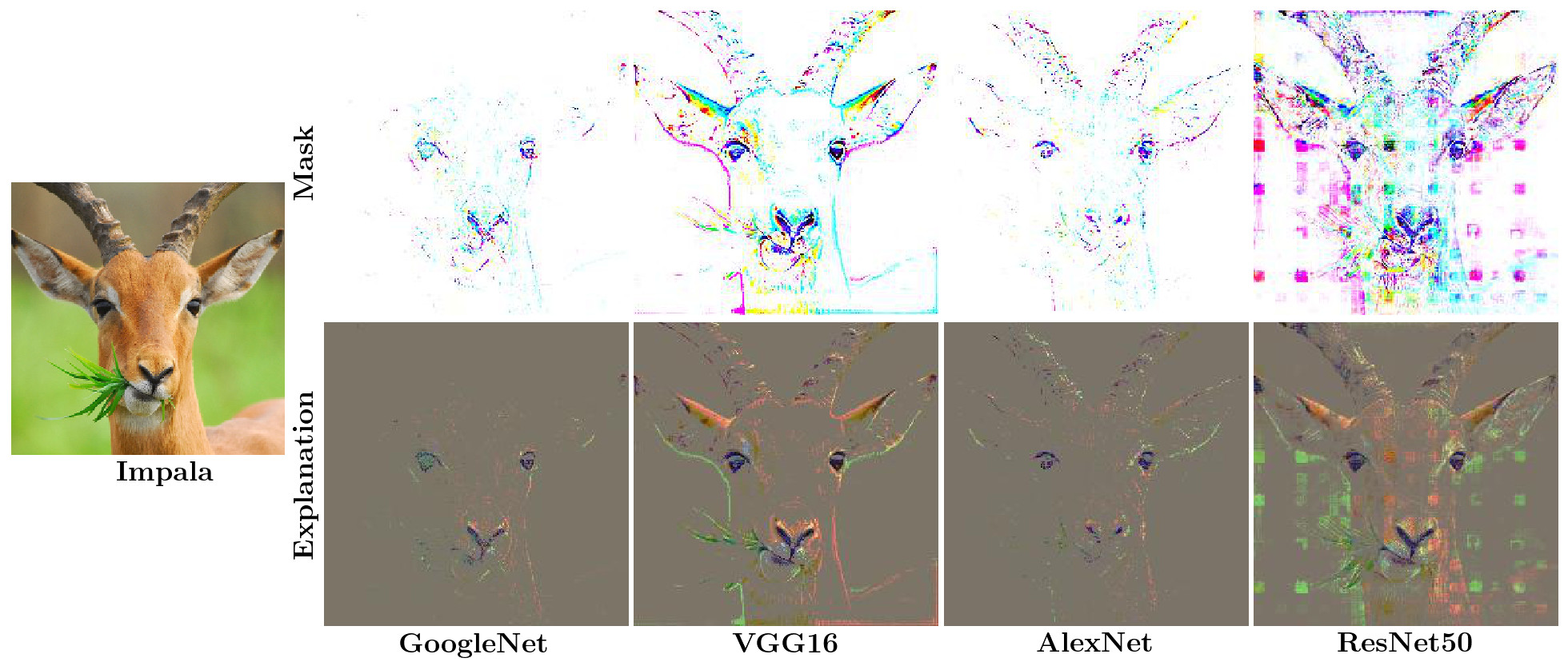}
	\caption{Masks and explanations computed using the \textit{deletion game} for different networks. }
	\label{fig_apx:impala}
\end{figure*}

\begin{figure*}[h!]
	\centering
	\includegraphics[width=0.95\linewidth]{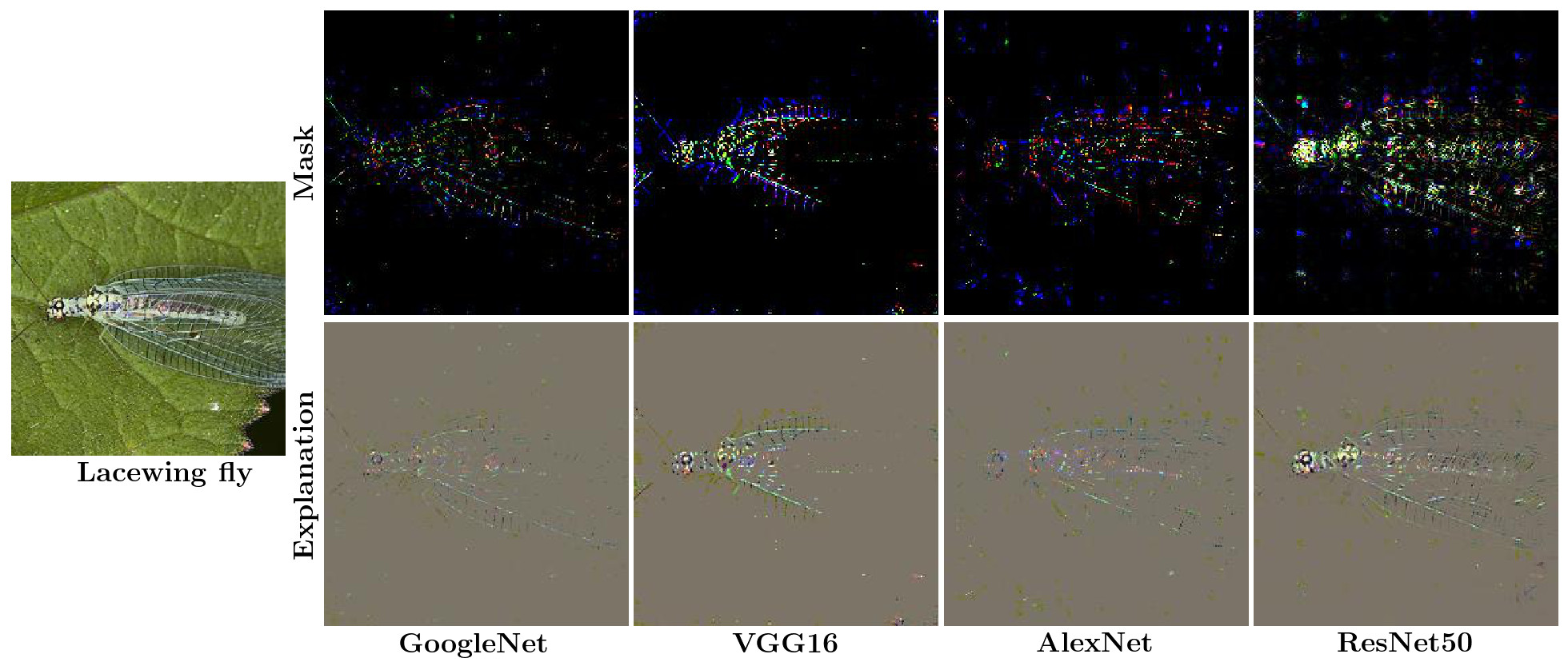}
	\caption{Masks and explanations computed using the \textit{preservation game} for different networks. }
	\label{fig_apx:lacewing}
\end{figure*}

\begin{figure*}[h!]
	\centering
	\includegraphics[width=0.95\linewidth]{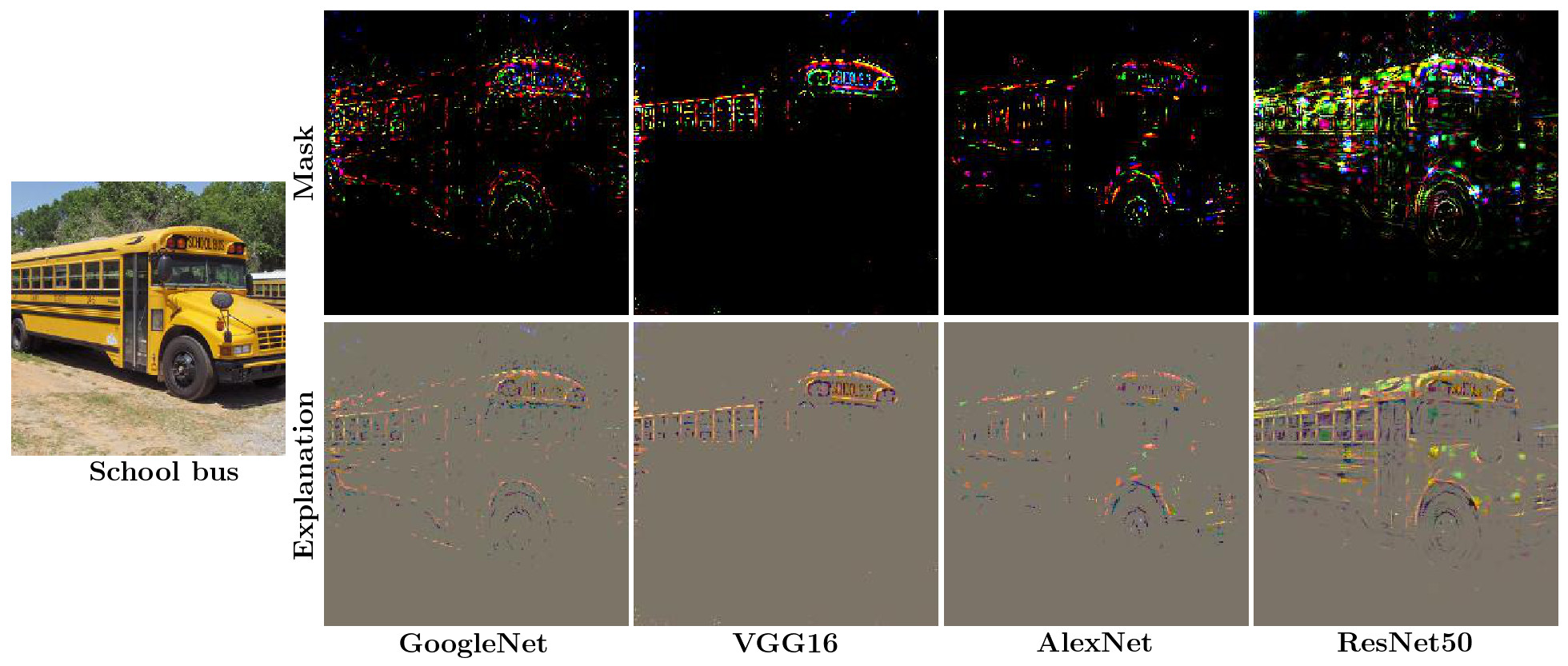}
	\caption{Masks and explanations computed using the \textit{preservation game} for different networks. }
	\label{fig_apx:schoolbus_Preservation}
\end{figure*}

In Fig.~\ref{fig_apx:lycaenid} and Fig.~\ref{fig_apx:impala} we compare the mask and explanation for four network architectures (\textit{GoogleNet, VGG16, AlexNet, ResNet50}) using the \textit{deletion game}. Respectively, in Fig.~\ref{fig_apx:lacewing} and Fig.~\ref{fig_apx:schoolbus_Preservation} we use the \textit{preservation game} for the same comparison. 

For all settings the explanations of \textit{ResNet50} and \textit{VGG16} are more dense, meaning more pixels have to be deleted/preserved to change/preserve the class prediction. This could be an indicator that these models are more robust, though, a detailed explanation would require further research. Besides, the grid-like pattern for the explanations from \textit{ResNet50}, described in Sec.~\ref{sec:interpretability} are visible. 

The importance of the color to classify the school bus (described in~Sec.~\ref{sec:biases}) can be seen in Fig.~\ref{fig_apx:schoolbus_Preservation}. 

For \textit{VGG16} we have observed that the pixels at the image edge are in many cases highlighted in the explanations. Furthermore, \textit{VGG16} shows pronounced edges in the explanation compared to the other networks. 

\subsection{Comparison of Games} 
\label{sec:appx_comparisonGames}
\begin{figure*}[h!]
	\includegraphics[width=\linewidth]{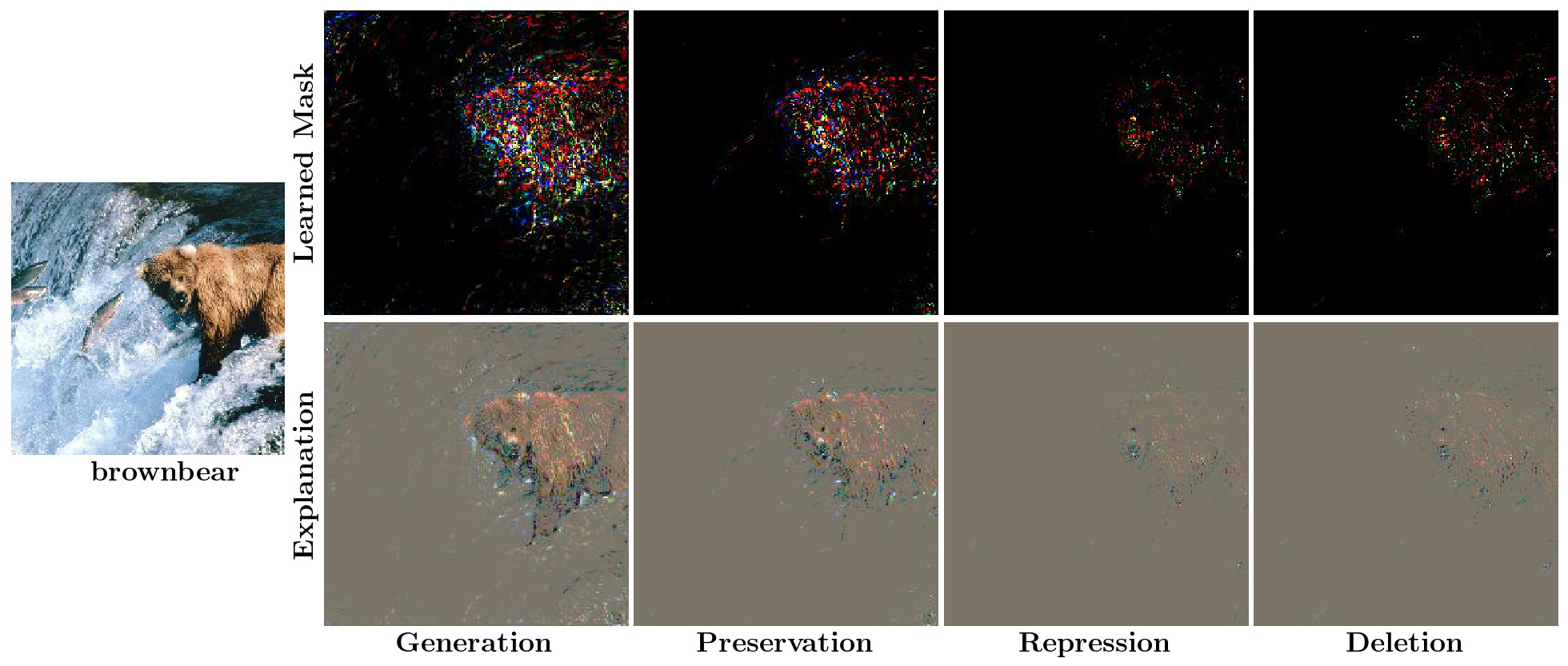}
	\caption{Explanations and masks computed using the different games for \textit{GoogleNet}. For the \textit{repression} and \textit{deletion game} the complementary masks ($1- \mathbf{m}$) are plotted to have true-color representations (see Sec.~\ref{sec:appx_comparisonNets}).}
	\label{fig_apx:brownbear}
\end{figure*}

\begin{figure*}[h!]
	\includegraphics[width=\linewidth]{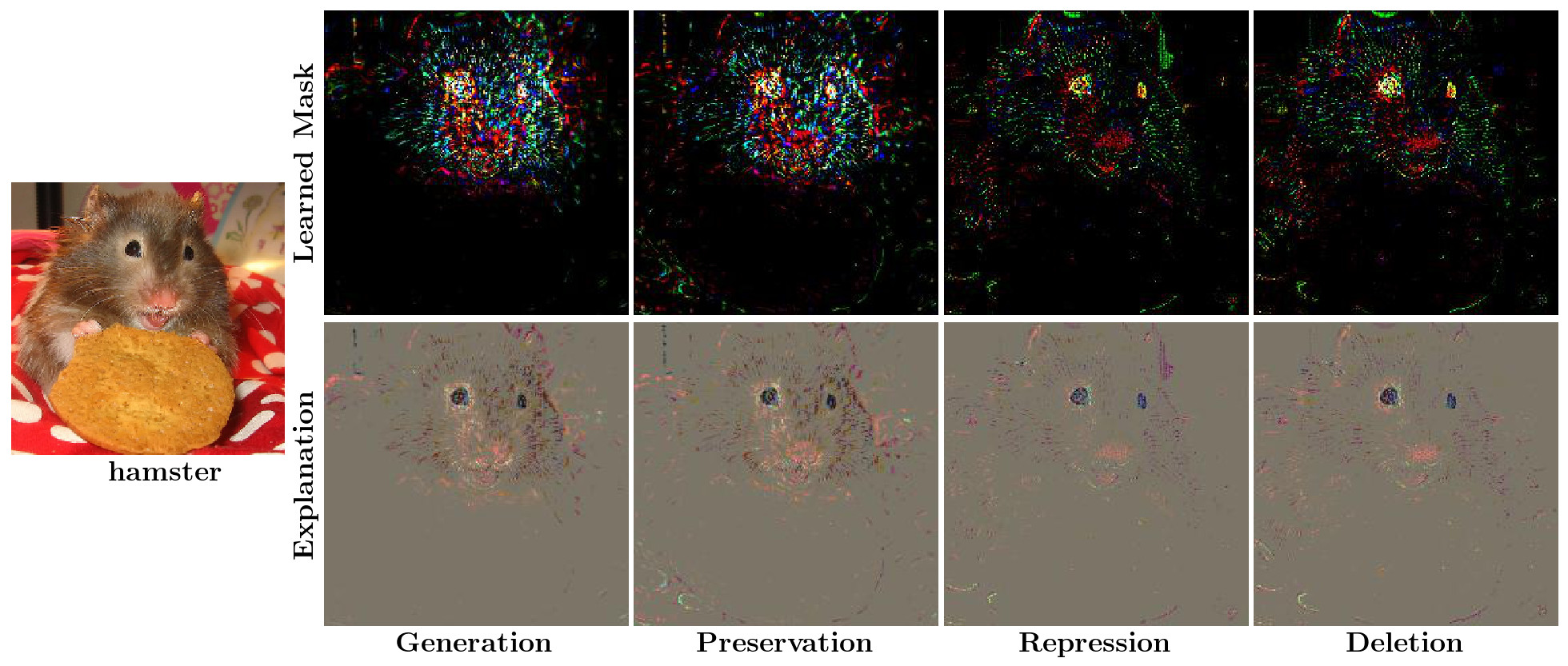}
	\caption{Explanations and masks computed using the different games for \textit{GoogleNet}. For the \textit{repression} and \textit{deletion game} the complementary masks ($1- \mathbf{m}$) are plotted to have true-color representations (see Sec.~\ref{sec:appx_comparisonNets}).}
	\label{fig_apx:hamster}
\end{figure*}

In Fig.~\ref{fig_apx:brownbear} and ~\ref{fig_apx:hamster} the different game types (see~Sec.~\ref{sec:perturbation_based_explanations}) are visually compared for \textit{GoogleNet}. 

The resulting explanations for the \textit{repression} and \textit{deletion game} are qualitatively similar. The similarity among the two games is due to both using the same optimization with only a different starting condition $\mathbf{m}=0$ for the \textit{repression} vs. $\mathbf{m}=1$ for the \textit{deletion game}. The same observation holds for the \textit{generation / preservation game.}  

The explanations of the \textit{repression} and \textit{deletion game} are more sparse compared to the \textit{generation / preservation game}. This is most likely due to the fact that only small parts of the image need to be suppressed to change the model output (\eg shifting one breed of dog to another), though, to evoke a certain model output one needs to create sufficient amount of evidence for this class. 

During the optimization only class pixels containing evidence towards the target class need to be changed for the \textit{generation} and \textit{deletion game}. After optimization most of the mask values stay zero for the \textit{generation game} and one for the \textit{deletion game}. The optimized masks are thus similar to its starting conditions. 

Vice versa, the opposite holds for the \textit{preservation} and \textit{repression game}. 

\subsection{Further Examples} \label{sec:appx_examples}

\begin{figure*}[h!]
	\centering
	\includegraphics[width=\linewidth]{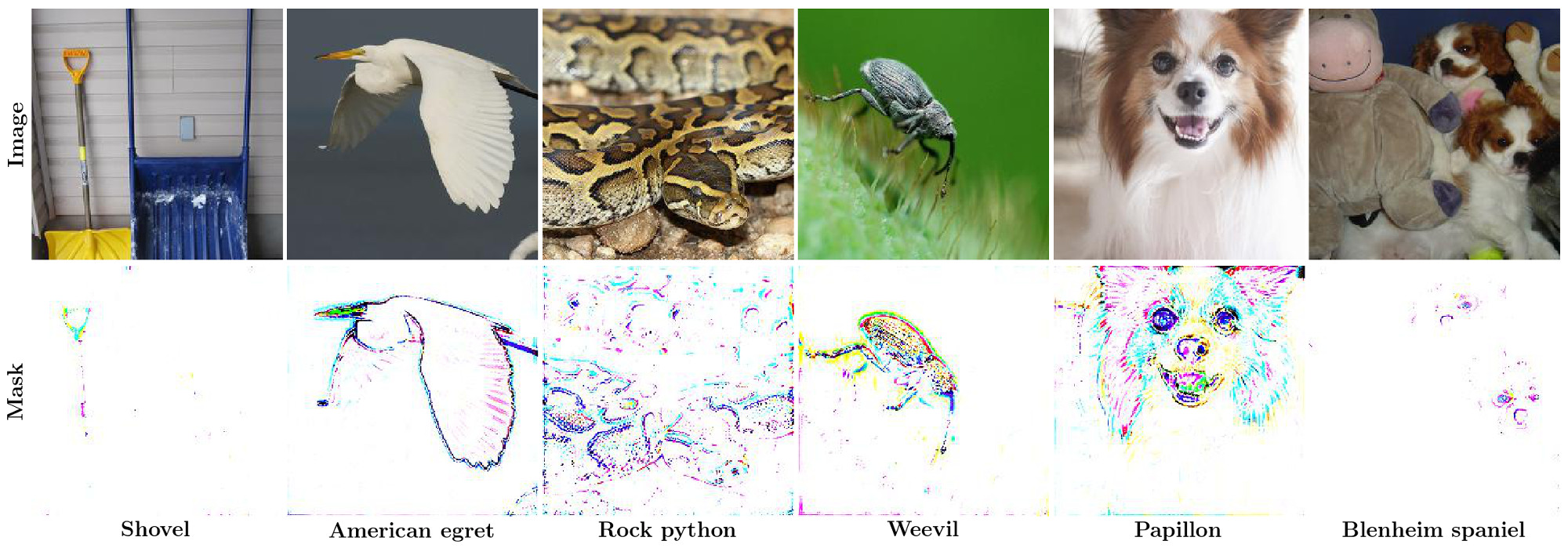}
	\caption{Explanation masks computed using the \textit{repression game} for \textit{VGG16}.}
	\label{fig_apx:vgg16_repression1}
\end{figure*}

\begin{figure*}[h!]
	\centering
	\includegraphics[width=\linewidth]{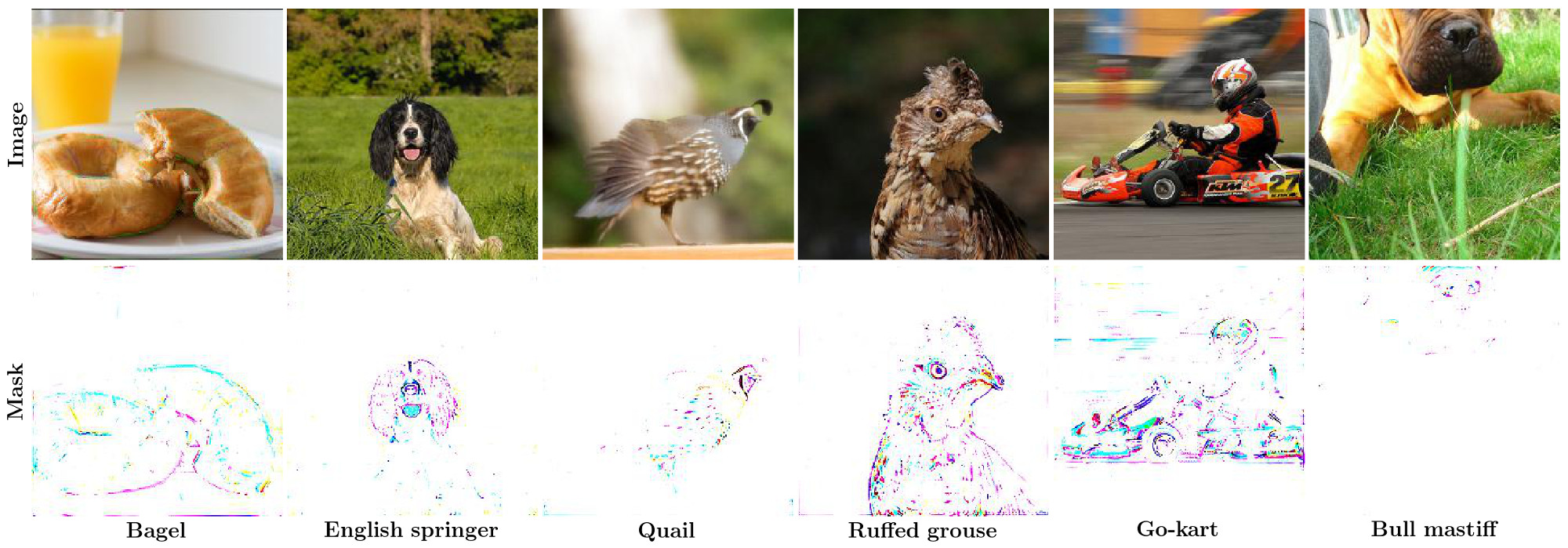}
	\caption{Explanation masks computed using the \textit{repression game} for \textit{VGG16}.}
	\label{fig_apx:vgg16_repression2}
\end{figure*}

\begin{figure*}[h!]
	\centering
	\includegraphics[width=\linewidth]{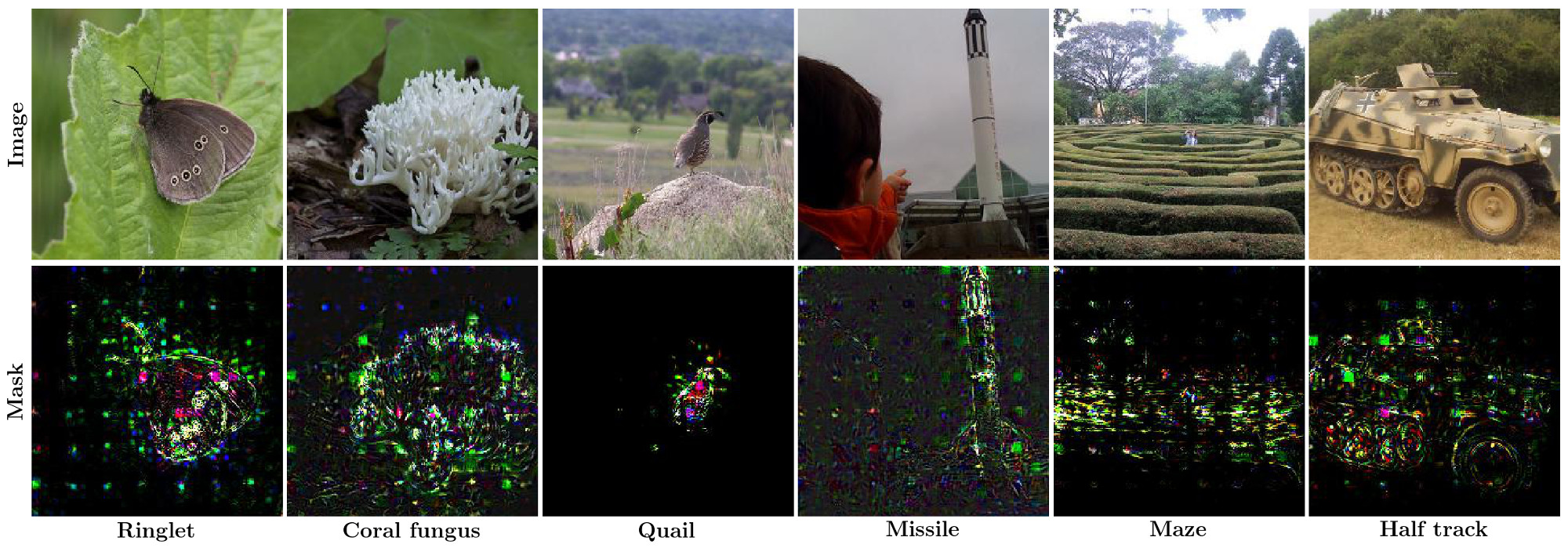}
	\caption{Explanation masks computed using the \textit{preservation game} for \textit{ResNet50}.}
	\label{fig_apx:resnet_preservation1}
\end{figure*}

\begin{figure*}[h!]
	\centering
	\includegraphics[width=\linewidth]{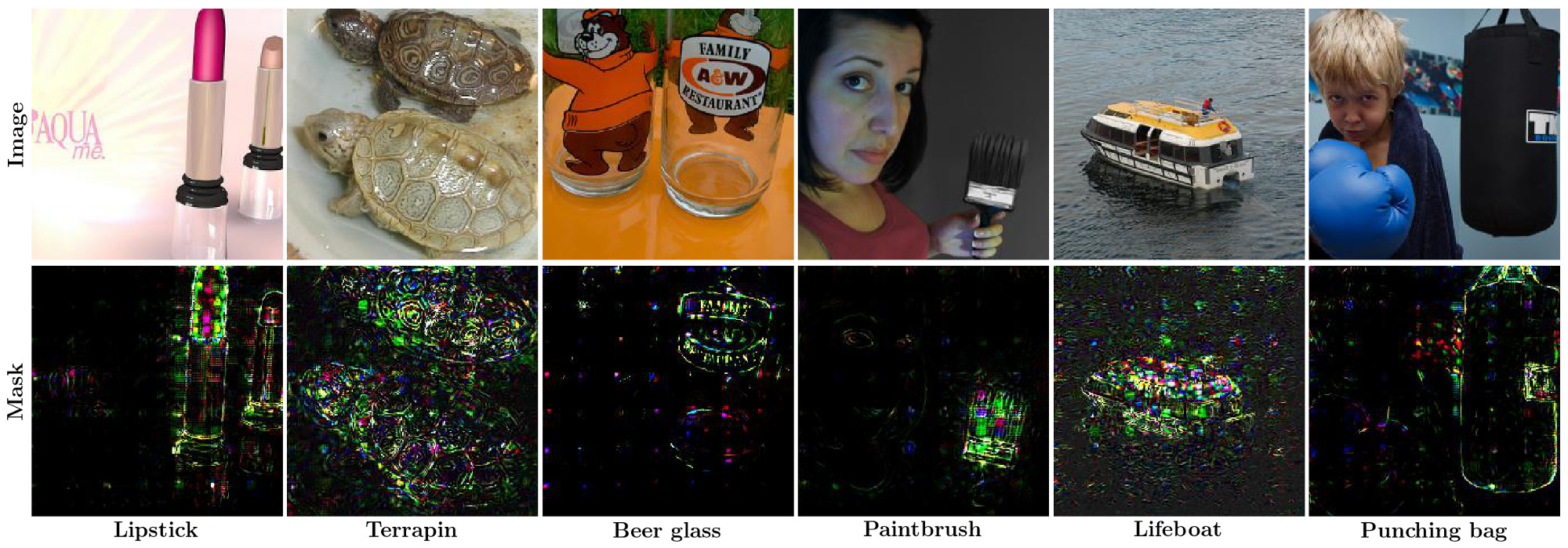}
	\caption{Explanation masks computed using the \textit{preservation game} for \textit{ResNet50}.}
	\label{fig_apx:resnet_preservation2}
\end{figure*}

In Fig.~\ref{fig_apx:vgg16_repression1},~\ref{fig_apx:vgg16_repression2},~\ref{fig_apx:resnet_preservation1}, and~\ref{fig_apx:resnet_preservation2} further explanations computed using FGVis are shown. 

\section{Quantitative Results} \label{sec:appx_quantitative_results}
\subsection{Faithfulness of Explanations} \label{sec:appx_faithfulnes}
To evaluate the faithfulness of our approach, we use the deletion metric of Petsiuk~\etal~\cite{petsiuk2018rise}. This metric measures how the removal of evidence affects the prediction of the used model.  The metric assumes that an importance map is given, which ranks all image pixels with respect to their evidence for the predicted class $c_{ml}$ (\ie the most-likely class). We use the mean mask (see~Sec.~\ref{sec:appx_comparisonGames}) as the pixel-wise importance map. The mean mask is computed for all images in the ImageNet validation dataset using the \textit{deletion game} with a learning rate of $0.3$ and a line-search to determine the $\lambda$ value. We iteratively use 4 equally spaced $\lambda$ values between $10^{-7}$ and $10^{-10}$ and stop when $y_{e}^{c_T} < 0.02 \cdot y_{x}^{c_T}$, where $y_{e}^{c_T}$ is the softmax score of class $c_T$ given the explanation and $y_{x}^{c_T}$ the corresponding score given the image.

\begin{figure*}[h!]
	\centering
	\begin{subfigure}{.23\linewidth}
		\centering
		\includegraphics[width=1.0\linewidth]{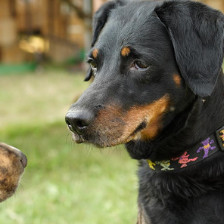}
		\caption{Image}
	\end{subfigure}%
	\begin{subfigure}{.515\linewidth}
		\centering
		\includegraphics[width=1.0\linewidth]{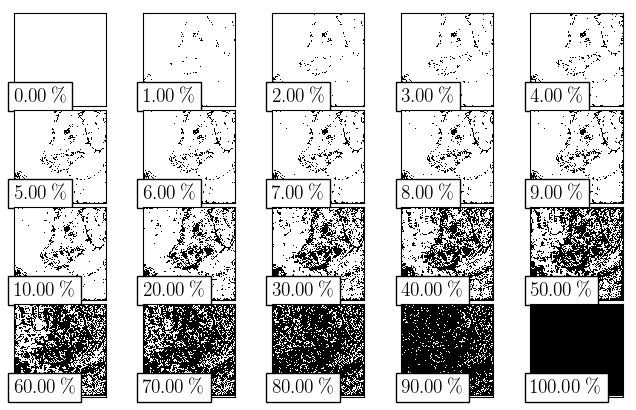}
		\caption{Binary deletion masks with fraction of removed pixels}
		\label{fig_apx:faithfulness_masks}
	\end{subfigure}
	\begin{subfigure}{.25\linewidth}
		\centering
		\includegraphics[width=1.0\linewidth]{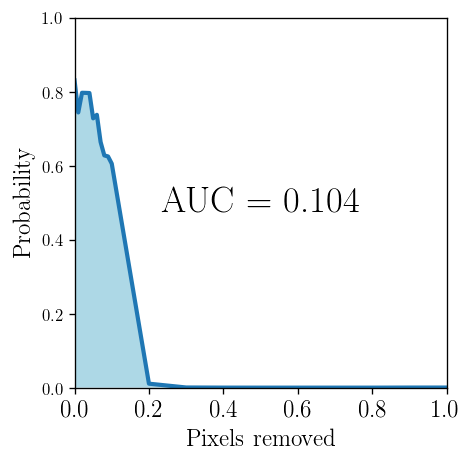}
		\caption{Deletion curve}
		\label{fig_apx:faithfulness_curve}
	\end{subfigure}
\caption{The deletion curve (c) is computed by successively deleting pixels (b) from the image according to their importance and measuring the resulting probability of the class $c_{ml}$.}
\label{fig_apx:faithfulness}
\end{figure*}

Using the importance map, the deletion curve is generated by successively removing pixels from the input image according to their importance and measuring the resulting probability of the class $c_{ml}$ (see Fig.~\ref{fig_apx:faithfulness_curve}). The removed pixels are set to zero, as proposed in Petsiuk~\etal~\cite{petsiuk2018rise}. The fraction of removed pixels is increased in increments of $0.25\%$ for the first $100$ steps and in increments of $1\%$ for the remaining $75$ steps. In Fig.~\ref{fig_apx:faithfulness_masks}, we visualize for an example image the binary masks used to successively set pixels to zero. For a clearer illustration, we reduced the number of deletion steps in this figure. The deletion metric is computed by measuring the \textit{area under the curve} AUC of the deletion curve (see Fig.~\ref{fig_apx:faithfulness_curve}) using the trapezoidal rule.

\subsection{Visual Explanation for Medical Images} \label{sec:retina_appx}
\textbf{Background of the disease:} As people with diabetes have a high prevalence for RDR \cite{yau2012global}, a frequent retinal screening is recommended and deep learning algorithms have been successfully developed to classify fundus images (\cite{colas2016deep}, \cite{gulshan2016development}, \cite{arunkumar2016multi}, \cite{ting2017development}). The black box character of these algorithms can be reduced by visual explanation techniques as shown in \cite{gondal2017weakly}. 

Of the publicly available 88,702 images \cite{kaggle_challenge} from EyePACS \cite{cuadros2009eyepacs}, we us 80\% for training and 20\% for validation for a classifier with binary outcome (referable diabetic retinopathy (RDR) vs. non-RDR) which is later used for the weakly-supervised localization. We use a similar setup as in \cite{gondal2017weakly} to train the binary classifier (RDR vs. non-RDR). 

Training was conducted with the same implementation settings as described in~\cite{gondal2017weakly} using an adopted version of the CNN architecture proposed by \cite{kaggle_oo} for classifying retinal images. We use leaky ReLUs as non-linearities and include batch normalization. 

The DiaretDB1 dataset \cite{kauppi2007diaretdb1} used to evaluate the weakly-supervised localization is a dataset of 89 color fundus images collected at the Kuopio University Hospital, Finland. All images have a resolution of 1500x1152 pixels and are scaled to the input dimension of the model. 

The dataset is ground truth marked by four experts. As proposed in \cite{kauppi2007diaretdb1} we consider pixels as lesions if at least three experts have agreed. 

We use FGVis with a fixed $\lambda=10^{-10}$ and a learning rate of $0.25$ stopping if the softmax score for RDR falls below 10\% with a maximum of 500 iterations.

\begin{figure*}[h!]
	\includegraphics[width=\linewidth]{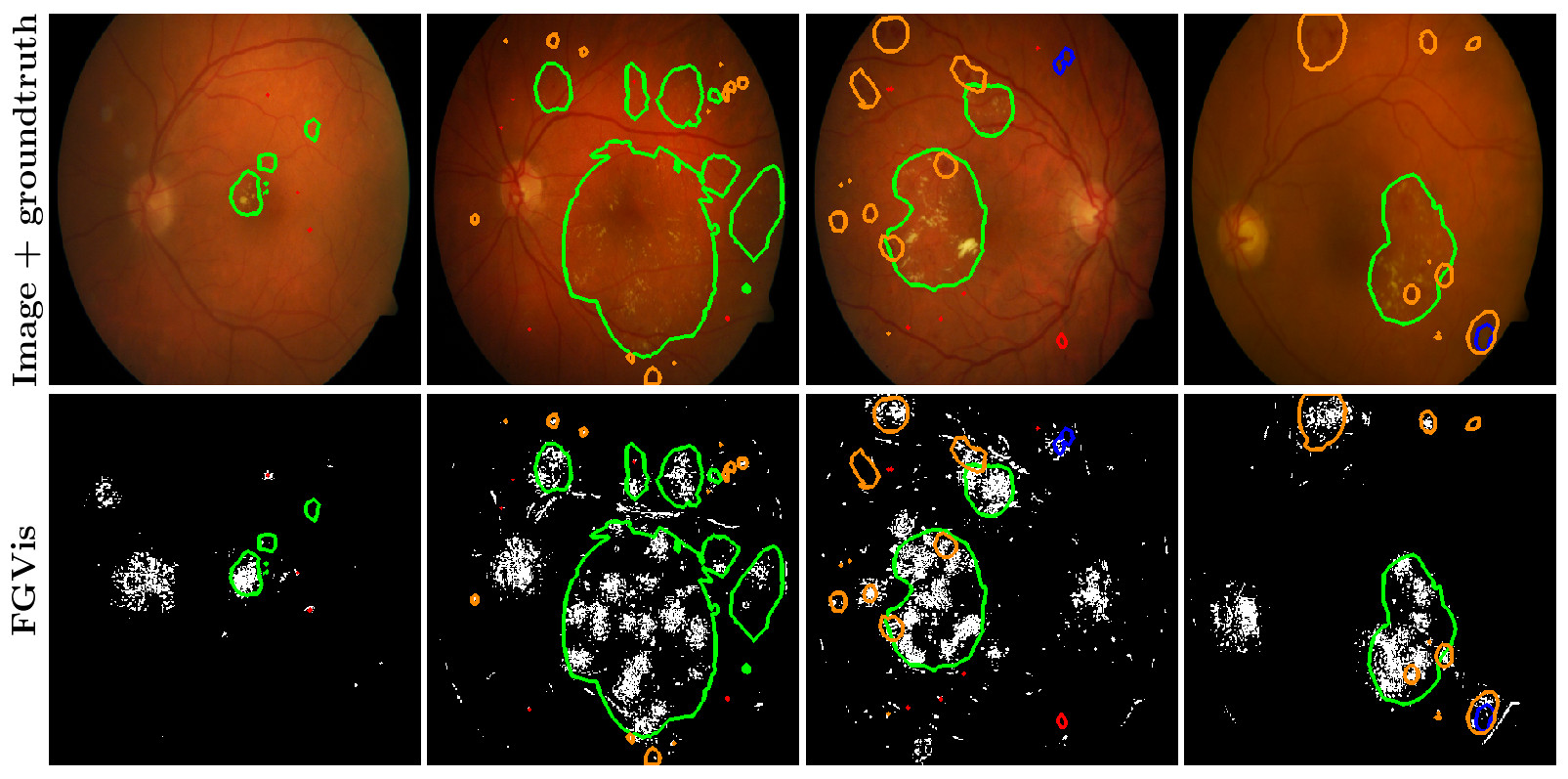}
	\caption{Weakly-supervised localization results on DiaretDB1 images. The top row shows fundus images, the bottom row our detection. All images are overlaid with ground truth markings in green (hard exudates), blue (soft exudates), orange (hemorrhages), red (red small dots). Though the network was trained in a weakly-supervised way given only the image label, most of the regions highlighted by FGVis fall within the ground truth markings. Note that mutations in the optic disk or blood vessels are an indicator for the disease~\cite{solomon2017diabetic} but these are not covered by the ground truth markings leading visually to false positives. FGVis highlights part of the blood vessels and optic disks. }
	\label{fig:diaretdb1}
\end{figure*}

In Fig.~\ref{fig:diaretdb1} retinal images overlaid with the ground truth (top row) are compared to our prediction (bottom row). 
To be consistent with~\cite{gondal2017weakly} the masks $m$ are binarized for better visualization and to be able to quantitatively report the sensitivity (see Tab.~\ref{tab:retina_img_level_se}).
Values greater or equal than 4\% of the maximum are set to one, the remaining pixels to zero.
The predicted pixels in the fine-grained masks $m$ map to the ground truth. Note that FGVis detects these pixels as they are the important ones to be deleted to reduce the softmax score for RDR. 

A medical expert would also look at mutations in the optic disk or blood vessels which additionally are an indicator for the disease~\cite{solomon2017diabetic}. These mutations are also highlighted by our method. They are not labelled in the ground truth markings leading to visual false positives (FPs). 

The strength of FGVis to visualize fine-grained structures can be seen in the detection of red small dots (microaneurysm) which are the earliest sign of diabetic retinopathy~\cite{agrawal2013survey}. As these often merely cover some pixels in the image, it is hard to detect them (zooming in~Fig.~\ref{fig:diaretdb1} is necessary to spot these). 